%% file: main.tex
\documentclass[10pt,twocolumn,letterpaper]{article}

\usepackage{3dv}
\usepackage{times}
\usepackage{epsfig}
\usepackage{graphicx}
\usepackage{amsmath}
\usepackage{amssymb}
\usepackage{mynotation}
\usepackage{algorithm}
\usepackage[noend]{algpseudocode}
\usepackage{graphicx}
\usepackage{booktabs}
\usepackage{pifont}
\usepackage{xcolor}
\usepackage{siunitx}
\usepackage{numprint}

\newcommand{\parsection}[1]{\vspace{0.5mm}\noindent\textbf{#1:}~}
\graphicspath{{./figures/}}

\usepackage[pagebackref=true,breaklinks=true,letterpaper=true,colorlinks,bookmarks=false]{hyperref}

\threedvfinalcopy 

\def\threedvPaperID{119} 

\ifthreedvfinal\fi

\begin{document}
	
	\title{Registration Loss Learning for Deep Probabilistic Point Set Registration}
	
	\author{Felix J\"aremo Lawin and Per-Erik Forss\'en \\
		\small Computer Vision Laboratory, Department of Electrical Engineering, Link\"oping University, Sweden\\
		\small\{\texttt{felix.jaremo-lawin},\; \texttt{per-erik.forssen}\}\texttt{@liu.se}
		{}
	}
	\maketitle
	
\begin{abstract}
Probabilistic methods for point set registration have interesting
theoretical properties, such as linear complexity in the number of
used points, and they easily generalize to joint registration of
multiple point sets. 
In this work, we improve their recognition performance to match state of the art.
This is done by incorporating 
learned features, by adding a von Mises-Fisher feature model in each
mixture component, and by using learned attention weights.
We learn these jointly using a registration loss learning strategy
(RLL) that directly uses the registration error as a loss, by back-propagating
through the registration iterations. This is possible as the
probabilistic registration is fully differentiable, and the result is
a learning framework that is truly end-to-end. 
We perform extensive experiments on the 3DMatch and Kitti datasets. The experiments demonstrate that our approach benefits significantly from the integration of the learned features and our learning strategy, outperforming the state-of-the-art on Kitti. Code is available at \href{https://github.com/felja633/RLLReg}{https://github.com/felja633/RLLReg}.
\end{abstract}
	
\input{introduction}

\input{related_work}

\input{method}
\input{experiments}
\input{conclusion}

\noindent\textbf{Acknowledgments}: This work was supported by the
\mbox{ELLIIT} Excellence Center and the Vinnova through the Visual
Sweden network 2019-02261.

{\small
	\bibliographystyle{ieee}
	\bibliography{references}
}

\setcounter{equation}{0}
\setcounter{figure}{0}
\setcounter{table}{0}
\setcounter{section}{0}
\renewcommand{\theequation}{S\arabic{equation}}
\renewcommand{\thefigure}{S\arabic{figure}}
\renewcommand{\thetable}{S\arabic{table}}
\renewcommand{\thesection}{S\arabic{section}}
\begin{center}
	\textbf{\large Supplementary Material}
\end{center}
\section{Introduction}

This supplementary material contains additional information that did
not fit into the main paper.
A reference implementation is available at \href{https://github.com/felja633/RLLReg}{https://github.com/felja633/RLLReg}.

\input{derivations}
\input{runtime}
\input{qualitative}
\newpage
\input{experiments_supp}
\end{document}

%% file: introduction.tex
\section{Introduction}

Point set registration is a fundamental computer vision problem, that has applications in 3D mapping and scene understanding. The task is to find the relative geometric transformations between scene observations, represented as unstructured point samples. There is a huge amount of different ways to approach point set registration, useful overviews can be found in e.g.~\cite{zhu19, choy2020deep}.

This work extends the paradigm of probabilistic point set registration \cite{GMMregPAMI11,evangelidis14,myronenko10}. In this paradigm, the scene is represented as a {\it Gaussian Mixture Model} (GMM) of the point set density.
This formulation has interesting theoretical properties, such as linear complexity in the number of used points, and it easily generalizes to joint registration of multiple point sets \cite{evangelidis14}. The recent top-performing methods in registration benchmarks, however, all use learned features \cite{choy2020deep,li2020}. To improve the probabilistic methods, we thus extend them to also benefit from learned features. To this end, we add a von Mises-Fisher feature model in each mixture component to model the local feature distribution.

In addition, this work also introduces a framework for true end-to-end training of feature descriptors in a single training phase. This is possible as the probabilistic registration
is fully differentiable. Descriptor learning is normally done in the contrastive paradigm, with a loss such as the triplet loss \cite{schroff2015} or its variants \cite{FCGF2019}, that encourages similarity of matches, while reducing similarity of mismatches.
The proposed {\it registration loss learning} (RLL) instead directly uses the registration error as a loss, by back-propagating through the registration iterations. This idea was also used in PointNetLK \cite{yaoki2019pointnetlk}, to learn a global descriptor vector. Here, we instead learn {\it local features} that are well suited to perform registration, also under partial overlap and varying point density. Using the registration error as a loss also provides an easy way to generate test data, by simply perturbing known registrations. Each training batch is thus composed of registration trajectories estimated using the current descriptor weights. The proposed formulation further allows us to perform training and testing on an arbitrary number of point sets jointly, making our approach easily adaptable to different applications. 

\parsection{Contributions} Our contributions are summarized as follows: \textbf{(i)} We extend probabilistic point set registration with a feature model, that can accomodate powerful deep features.
\textbf{(ii)} The features are learned using {\it registration loss learning} (RLL), by back-propagating gradients through EM iterations.
To avoid instability during training, we propose a robust loss function that is insensitive to large registration errors. 
\textbf{(iii)} As RLL is end-to-end, it allows simultaneous learning of pointwise attention weights, that focus registration on important regions.

\begin{figure*}
	\centering
	\includegraphics[width=0.98\textwidth, trim={0 102mm 0 0},clip]{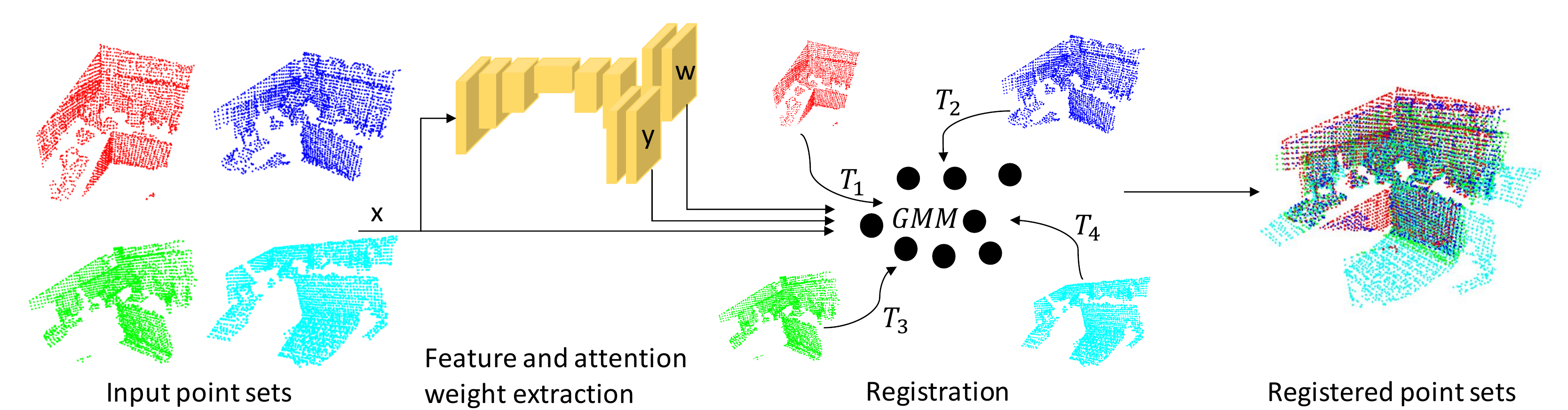}
	\caption{Overview of our registration algorithm. First, a CNN extracts features, $y$, and attention weights, $w$, from the input point sets, $x$. The points and features are then modeled using a GMM with a von Mises-Fisher distribution for each component to model the features. Finally, the parameters of the GMM, the feature model, and the transformations $T_i$ are jointly refined until convergence.}
	\label{fig:overview}
\end{figure*}

Figure \ref{fig:overview} gives an overview of the proposed approach.
It is tested by extensive experiments on the 3DMatch and Kitti odometry datasets and an ablative analysis highlights the impact of the proposed components. The experiments demonstrate that our approach benefits significantly from the integration of the feature model and outperforms the state-of-the-art on Kitti.

%% file: related_work.tex
\section{Related work}
Point set registration has received a high interest over the years. In this section, we outline related works categorized as classical, feature matching, and probabilistic methods.

\parsection{Classical} The iterative closest point (ICP) based methods~\cite{chen91,besl92,pomerlau15}, iteratively find putative correspondences given the current transformation parameters through a nearest neighbor search in the spatial domain. While being efficient, the ICP methods rely heavily on a good initialization of the transformation parameters. More robust variants of ICP have been proposed, employing multiple restarts~\cite{ICP_restarts}, graph optimization~\cite{theiler2015globally}, or branch and bound~\cite{Campbell_2016_CVPR} to find globally optimal transformation parameters.

Another branch of methods are based on point neighborhood density~\cite{tsin2004correlation}. These methods find the relative transformation between point sets by modeling them as densities using a kernel, and then minimize a loss based on correlation or divergence~\cite{GMMregPAMI11}. The density based losses have improved monotonicity compared to the $L^2$ loss used in ICP~\cite{tavares2020assessing}.   

\parsection{Feature matching} In order to increase the accuracy of the correspondence set, it is beneficial to use geometric features~\cite{rusu09}. Such methods generally employ RANSAC based frameworks to obtain the transformation parameters given putative feature correspondences. The FGR method \cite{zhou2016fast} employs geometric features and iteratively optimizes the correspondence set by minimizing a robust loss. 

Since matching approaches are reliant on the discriminative properties of the features, more recent approaches~\cite{yew2018-3dfeatnet, deng2018ppfnet,deng2018ppfnet,FCGF2019,deng20193d, gojcic2020learning, choy2020deep} aim at learning these using deep networks. The networks are generally based on e.g.~PointNet~\cite{qi2016pointnet} or sparse 3D convolutions~\cite{choy20194d} and are usually trained using contrastive learning techniques. With these features, registration can be performed by first establishing correspondences through feature matching and then employing a RANSAC scheme to obtain a robust estimation~\cite{yew2018-3dfeatnet, deng2018ppfnet,deng2018ppfnet,FCGF2019,deng20193d,li2020}. The RANSAC step was replaced in recent approaches~\cite{gojcic2020learning, choy2020deep} by further learning an importance weighting of the correspondences. A robust estimate of the transformation parameters can then be found using a weighted Procrustes solver, which in contrast to RANSAC is fully differentiable. Consequently, these methods can benefit from end-to-end learning using the registration objective as a loss. The work in~\cite{gojcic2020learning} additionally proposes an approach for joint registration of multiple point sets using an iterative refinement of pairwise registrations.

The works in \cite{wang2019deep, wang2019prnet, lu2019deepvcp} extend ICP by employing features in the correspondence establishment step. In~\cite{yaoki2019pointnetlk}, 
a distance is minimized, between global representations of the points using a gradient decent optimization. While showing promising results on synthetic data, the methods in~\cite{wang2019deep, wang2019prnet, yaoki2019pointnetlk} fail to generalize to real world point sets~\cite{choy2020deep}.

\parsection{Probabilistic} Probabilistic approaches jointly model the underlying point set distribution and infer the transformations. As a result, the putative correspondences become probabilistic, leading to less sensitivity to measurement noise and the initial transformation. In the Coherent Point Drift (CPD) approach~\cite{myronenko10}, the source point cloud is modeled as a GMM, conditioned on the target point set. The parameters of the GMM and the transformation parameters are then optimized using expectation maximization. A downside of the CPD approach is its high computational complexity. To remedy this, recent approaches employ hierarchical GMMs~\cite{eckart2018hgmr} or a fast filtering formulation~\cite{gao18filterreg} to achieve similar performance at significantly higher speed. The CPD approach was generalized in~\cite{evangelidis14, evangelidis18} to multi-view registration by also optimizing the GMM cluster centers instead of fixing them to the source point set. In~\cite{jaremo18a}, a point set density weighting was proposed for the probabilistic methods to handle density variations in the point set.

Feature models were incorporated in the probabilistic methods to increase robustness in~\cite{DanelljanICPR2016, DanelljanCVPR2016}. These models, however, were based on color channels and handcrafted geometric features. In this work we aim to bridge the gap between feature matching based and probabilistic approaches by learning both point features and attention weights.

%% file: method.tex
\section{Method}
We propose a probabilistic method for point set registration, which exploits the powers of learned features. 
Based on the generative model in~\cite{evangelidis14}, we introduce a strategy for integrating and learning deep feature representations of the local point regions. Specifically, we model the joint distribution of points and features as a Gaussian Mixture Model (GMM), where each component represents the density of the spatial coordinates and features in a local region. Hence, instead of employing point matching techniques, correspondences are indirectly represented as soft assignments to the components. This enables a generalization to joint registration of an arbitrary number of point sets. Moreover, the computational complexity is linear with the number of mixture components and points. This allows us to significantly reduce the complexity in comparison to previous deep approaches by keeping the number of components low. Finally, registration is performed by jointly optimizing the parameters of the GMM and the transformation parameters using Expectation Maximization (EM). The EM iterations are both efficient and fully differentiable, allowing us to learn features for registration in an end-to-end manner by minimizing a loss based on the registration error. 

Our registration algorithm can be broken down into two steps, as illustrated in Figure.~\ref{fig:overview}; a feature extraction step, and a registration step. The feature extractor network
processes each point set separately and generates pointwise features and attention weights. These are incorporated into a probabilistic model, which is optimized with EM in the registration step to obtain the transformation parameters. In the following sections we
detail the approach. 

\subsection{Probabilistic Point Set Registration}
In this section we provide a generic description of probabilistic registration. Let $\mathcal{X}_i = \{x_{ij}\}_{j=1}^{N_i}, i = 1,\ldots, M$ be $M$ point sets, consisting of 3D-point observations $x_{ij} \in \reals^3$ independently drawn from the distributions $p_{X_i}$. We assume that these distributions are all related to a common scene distribution $p_V$  by the rigid 3D transformations $T_{\omega^i} : \reals^3 \rightarrow \reals^3$ parameterized by $\omega^i$ such that $p_V(T_{\omega^i}(x)|\theta)=p_{X_i}(x|\theta)$. The aim is to jointly infer $p_V(v|\theta)$, with parameters $\theta$, and the optimal transformation parameters for $i = 1, \ldots M$.

We model the density $p_V(v|\theta)$ as a mixture of Gaussian distributions,
\begin{equation}
\label{eq:GMM}
p_V(v|\theta) = \sum_{k=1}^K \pi_k \norm (v;\mu_k, \Sigma_k) \,.
\end{equation}
Here, $\norm (v;\mu, \Sigma)$ is a normal density function with mean $\mu$ and covariance $\Sigma$. The number of components is denoted by $K$ and $\pi_k$ is the mixing weight of component $k$ satisfying $\sum_k \pi_k=1$. In order to reduce the number of parameters in the mixture model, we consider fixed uniform mixing weights $\pi_k = 1/K$ and isotropic $\Sigma_k=\sigma_k^2{\bf I}$. We denote the set of all mixture parameters by $\theta = \{\pi_k, \mu_k, \Sigma_k\}_{k=1}^K$ and the set of all parameters in the model by $\Theta = \{\theta, \omega^1, \ldots, \omega^M\}$. For brevity, in the following sections we omit the parameters $\omega^i$ in the notation for the transformation functions, i.e~we set $T_i = T_{\omega^i}$.

\subsubsection{Inference}
\label{sec:inference}
To infer the parameters $\Theta$, we maximize the log-likelihood of the observed points $x_{ij}$:
\begin{equation}
\label{eq:ML}
\mathcal{L}(\Theta;\mathcal{X}_1,\ldots,\mathcal{X}_M) = \sum_{i}^M \sum_j^{N_i} \log (p_V(T_i (x_{ij})|\theta))\,.
\end{equation}
Maximization is done using EM, where we introduce the latent assignment variables $Z \in (1, \ldots, K)$, assigning points to the mixture components. Given the current parameter state $\Theta^n$, an EM iteration maximizes the expected complete data log-likelihood objective,
\begin{equation}
\label{eq:EM1}
\mathcal{Q}(\Theta;\Theta^n) =\sum_{i}^{M}\sum_j^{N_i}\!E_{Z|x,\Theta^n}\! \left[\log\left(p_{V,Z}(T_i (x_{ij}),Z|\theta)\right) \right] w_{ij} .
\end{equation}
Here, $w_{ij}$ is a pointwise attention weighting that determines the importance of each point in the objective. In~\cite{jaremo18a}, $w_{ij}$ was introduced as an inverse local density estimate to account for variations in the sample densities $p_{X_i}$. Another option which we will explore is to learn a predictor for $w_{ij}$. 
 
In the E-step, we evaluate the expectation in \eqref{eq:EM1}, taken over the probability distribution of the assignment variables,
\begin{align}
\label{eq:latent_posterior}
p_{Z|X_i}(k|x,\Theta) &= \frac{p_{X_i,Z}(x,k|\Theta)}{\sum_{l=1}^{K} p_{X_i,Z}(x,l|\Theta)} \nonumber\\
& = \frac{\pi_k \norm (T_i (x); \mu_k, \Sigma_k)}{\sum_{l=1}^K \pi_l \norm (T_i (x); \mu_l, \Sigma_l)} \,.
\end{align}

For compactness, we introduce the notation $\alpha_{ijk}^n = p_{Z|X_i}(k|x_{ij},\Theta^n)$. In the subsequent M-step, we maximize,
\begin{equation}
\label{eq:EM2}
Q(\Theta;\Theta^n) = \sum_{i=1}^{M} \sum_{j=1}^{N_i} \sum_{k=1}^{K} \alpha_{ijk}^n w_{ij} \log\left(p_{V,Z}(T_i (x_{ij}),k|\theta)\right) \,, 
\end{equation}
with respect to $\Theta$.

Due to the non-linearities of \eqref{eq:EM2}, a closed form solution to the joint maximization problem over $\theta$ and $T_i$ is unlikely to exist. As in \cite{evangelidis14,evangelidis18} and \cite{myronenko10}, we circumvent this by employing the {\it expectation conditional maximization} (ECM) method. In this approach, the transformations $T_i$ are found first, given the latent assignments and the mixture parameters $\theta$. Afterwards the mixture parameters are found given the latent assignments and the transformation parameters. As both sub-problems have closed-form solutions, this leads to an efficient optimization procedure. We refer to \cite{evangelidis14} for the derivations of $T_i$, $\mu_k$ and $\sigma_k$.

\subsection{Probabilistic Feature Representation}
The formulation in the previous section only considered the spatial coordinates of the points in the probabilistic registration model. In this section we extend the model to the joint distribution $(X,Y)\sim p_{X,Y}(x,y)$ over the spatial coordinates $x$ and features $y$ of the points. Specifically, we assume that the features are invariant to the transformations $T$. Further, as in~\cite{DanelljanCVPR2016}, we also assume that features and spatial locations are independent, conditioned on the assignment variable $Z$. We then maximize the likelihood for a model over the joint density function $p_{X,Y}(x,y|\Theta, \nu)$. Here, we have introduced feature model parameters $\nu$. The conditional independence between $X$ and $Y$, and the assumption that $Y$ is invariant to $\Theta$, allow us to factorize $p$ as,
\begin{align}
p(x,y,k|\Theta, \nu)&= p(x,y|k,\Theta, \nu) p(k|\Theta, \nu) \nonumber\\&= p(x|k,\Theta)p(k|\Theta, \nu)p(y|k,\nu)\,.
\end{align}
Again, we assume that the mixing weights $\pi_k = p(k|\Theta, \nu)$ are constant and equal for all components. Using the above factorization, the E-step turns into evaluation of the following posterior distribution,
\begin{align}
\label{eq:latent_posterior3}
&p_{Z|X_i,Y}(k|x,y,\Theta,\nu_k) = \frac{p_{X_i,Y,Z}(x,y,k|\Theta, \nu_k)}{\sum_{l=1}^{K} p_{X_i,Y,Z}(x,y,l|\Theta,\nu_l)} \nonumber\\ 
&= \frac{p_{X_i|Z}(x|k,\Theta)p_{Z}(k|\Theta)p_{Y|Z}(y|k,\nu_k)}{\sum_{l=1}^{K} p_{X_i|Z}(x|l,\Theta)p_{Z}(l|\Theta)p_{Y|Z}(y|l,\nu_l)} \nonumber\\
&= \frac{\pi_k \norm (T_i (x); \mu_k, \Sigma_k)p_{Y|Z}(y|k,\nu_k)}{\sum_{l=1}^K \pi_l \norm (T_i (x); \mu_l, \Sigma_l)p_{Y|Z}(y|l,\nu_l)} \,.
\end{align}

Note that, the terms in the expression are identical to the terms in \eqref{eq:latent_posterior}, except for the added $p_{Y|Z}(y|k,\nu_k)$. In the M-step, we maximize,
\begin{align}
\label{eq:obj2}
&F(\Theta;\Theta^n) = \notag \\
&\sum_{i=1}^{M} \sum_{j=1}^{N_i} \sum_{k=1}^{K} w_{ij}\alpha_{ijk}^n \log(p_{V,Y,Z}(T_i (x_{ij}), y_{ij},z=k|\theta, \nu_k))\nonumber\\
&= Q(\Theta;\Theta^n)+ \sum_{i=1}^{M} \sum_{j=1}^{N_i} \sum_{k=1}^{K} w_{ij}\alpha_{ijk}\log p_Y(y_{j}|z=k,\nu_k) \,.
\end{align}
Here, $Q$ is identical to the objective in the M-step i equation \eqref{eq:EM2} and only depends on $T_i$ and $\theta$. It can thus be maximized using the procedure in section \ref{sec:inference}. The second term contains the feature model and only depends on the feature parameters $\nu_k$. In total, we optimize the parameters of the original spatial distribution of the probabilistic model along with one feature parameter vector $\nu_k$ for each component.

\subsubsection{Feature Model}
\label{sec:feat}
We aim to integrate powerful deep features in the probabilistic model. Deep features are generally high dimensional, providing a discriminative representation of the local appearance of a given point. To be able to efficiently model these features and enable end-to-end learning, we propose a von Mises-Fisher distribution~\cite{mardia2000}, with parameters $\nu_k$ to represent the density $p_Y(y_{ij}|z=k,\nu_k)$. We therefore assume that the feature vectors are normalized, i.e.~$\big \| y \big \| = 1$.
Thus, in the E-step we compute,
\begin{equation}
p_{Y|Z}(y|k,\nu_k) \propto e^{\nu_k^T y_/s^2}
\end{equation}

In the M-step, we update $\nu_k$ by solving
\begin{equation}
\max_{\nu_k} \sum_{i=1}^{M} \sum_{j=1}^{N} w_{ij} \alpha_{ijk}\nu_k^T y_{ij}\,,\;\;\text{subject to }\|\nu_k\|=1\,.\end{equation}
The solution to the above maximization problem is
\begin{equation}
\nu_k =\frac{\sum_{i=1}^{M} \sum_{j=1}^{N} w_{ij}\alpha_{ijk} y_{ij}}{\big \| \sum_{i=1}^{M} \sum_{j=1}^{N} w_{ij}\alpha_{ijk}y_{ij} \big\|}\,.
\label{eq:ml-von}
\end{equation}
We provide a derivation of \eqref{eq:ml-von} in the supplementary material. Note that, $s$ can be considered as a parameter to be optimized~\cite{mardia2000}. In this work, however, we treat it as a hyper-parameter kept fixed during inference. In our experiments, we set $s=0.4$.

\subsubsection{Feature Extractor}
We construct a feature extractor $F_{\phi}$, parameterized by $\phi$. The feature extractor takes 3D coordinates of a point set as input and outputs features $y$ and attention weights $w$ for each point. Similar to the recent works~\cite{gojcic2020learning, choy2020deep}, we base our network on the sparse 3D conv-net in~\cite{FCGF2019}. This network comprises four encoder blocks and decoder blocks which are connected in a Unet-based structure. Each block consists of a conv-layer, a ReLU activation, and a batch normalization layer. 

On top of the final block, our network generates feature maps $y$, and attention weight maps $w$ in two separate streams. Both streams consist of two conv-layers with a ReLU in between. The output of the feature stream is $L^2$-normalized to fit the distribution of our probabilistic feature model in section \ref{sec:feat}. In the attention stream we produce a scalar that is input to a SoftPlus activation. This ensures both positive weighting and large flexibility. 

\subsubsection{Registration Algorithm}
In this section we outline the details of our registration algorithm. For a given set of $M$ overlapping point sets $\left\{\mathcal{X}_i\right\}_1^M$, we seek the unknown relative transformations $T_{ij}$. 

First, we apply the feature extractor $F_{\phi}(\mathcal{X}_i)$ on the point sets $\mathcal{X}_i, i = 1,\ldots, M$ individually, to obtain features $\{y_{ij}\}_{j=1}^{N_i}$ and attention weights $\{w_{ij}\}_{j=1}^{N_i}$. Next, we employ our probabilistic registration algorithm. When starting from an unknown GMM, its parameters need to be initialized. We initialize the means $\mu_k$ of the spatial mixture components by randomly sampling points on a sphere centered at the mean of the point sets, and with a radius equal to the point sets standard deviation. Further, we initialize the standard deviations $\sigma_k$ of the mixture components as the maximum distance between the points. We assume that the feature distribution is uniform in the first iteration, thus it has no impact on the first transformation estimate. Next, we apply the EM algorithm by iterating the E and M-steps (see Sections \ref{sec:inference} to \ref{sec:feat}). In order to reduce the sensitivity to the initialization step, we keep $\mu_k$ fixed during the first two iterations. After $N_\text{iter}$ iterations, we obtain the final transformation estimates $\left\{T_{i}\right\}$, which map the point sets from their local coordinates to the GMM frame. 

\subsection{Registration Loss Learning}
\label{sec:training}
We learn the parameters $\phi$ of the feature extractor by minimizing a registration loss over a dataset $\mathcal{D} \subset 2^{\displaystyle \left\{\mathcal{X}_i\right\}_1^M}$, consisting of $D$ groups $\mathcal{D}_d\in\mathcal{D}$ of overlapping point sets. This loss can be written as
\begin{equation}
\mathcal{L}(\phi; \mathcal{D})= \sum_{d=1}^D \mathcal{R}(\phi; \mathcal{D}_d)\,.
\label{eq:loss}
\end{equation}
We denote the number of point sets in each group by $M_d=|\mathcal{D}_d|$. In principle, $M_d$ can be an arbitrary number of point sets, but on order to reduce training speed and memory consumption, we fix it to two or four sets in our experiments. We further assume that the ground truth relative transformations $T^\text{gt}_{ik}$ from point set $k$ to $i$ are known in the training set. Similar to~\cite{gojcic2020learning}, the sample loss $\mathcal{R}$ is computed over pointwise errors given the estimated relative transformations $T^n_{ik}=(T^n_i)^{-1} \circ T^n_k$ for each EM iteration $n$ as
\begin{align}
&\mathcal{R}(\phi; \!\mathcal{D}_d)\! =\nonumber\\ &\sum^{N_\text{iter}}_n \!v_n\!\!\!\!\sum^{M_d-1,M_d}_{i=1,k=i+1}\! \frac{1}{N_i}\!\sum^{N_i}_{j=1}\!\rho(\big\|T^n_{ik}(x_{ij})- T^\text{gt}_{ik}(x_{ij})\big\|_2/c)\!\,.
\label{eq:sample_loss}\end{align} 
The estimated transformations $T^n_i$ and $T^n_k$ are found using the proposed EM approach.
We observed that large errors in the the registration loss causes instability during training.
To counter this we employ a Geman-McClure penalty $\rho$. This is applied to the Euclidean distance between the points mapped by the estimated transformation and the ground truth transformation. 
The parameter $c$ is a scale factor that determines the steepness of the error function. Crucially, this robust penalty function ensures that the loss is not dominated by samples far from the solution. To further counter the influence of examples far from convergence, the loss is weighted with $v_n = 1/(40-n)$, such that the later iterates will have a larger impact. 

As all steps in the EM algorithm are differentiable, we can update the parameters $\phi$ by back-propagating the errors of the loss in \eqref{eq:loss} through the registration procedure. The feature extractor can thus be trained end-to-end to maximize registration accuracy.

\parsection{Details} We train our approach on datasets of overlapping point set pairs using the ADAM optimizer with initial learning rate of $0.004$ and a batch size of 6. We split the training into 180 epochs with 2000 samples each. Every 40th epoch we reduce the learning rate by a factor of $0.2$. In order to increase variations in the dataset we apply random rotations and translations to the point sets. Specifically, to construct the rotations we draw random rotation axes and rotate with an angle between 0 and $\pi/8$ radians. The random translation vectors are drawn from a uniform distribution and have a random length depending on the dataset. For large scale lidar datasets we draw translation vectors with a maximum norm of $2$ meters while for smaller scale RGB-D datasets the maximum length is set to $0.8$ meters. During training we set the number of mixture components $K=50$. Further, we found in our experiments that using $N_\text{iter}=23$ EM iterations gives good results (see Section \ref{sec:ablation}).   

%% file: experiments.tex
\section{Experiments}
\label{sec:exp}
We evaluate our approach on the RGB-D sequences from the 3DMatch datset~\cite{zeng20163dmatch} and the lidar point clouds from the Kitti odometry dataset~\cite{Geiger2012CVPR}. 3DMatch contains sequences from 62 scenes, 54 of which are for training and 8 for testing. We
use 7 sequences from the training set for
validation and hyper-parameter tuning. Kitti contains scans from 11 scenes (00-10), including 8 for training and 3 for testing. We use (00-05) for training and (06-07) for validation. 

In all experiments, we collect samples by first drawing a reference frame from a sequence. We then randomly sample frames that are nearby, in terms of both frame id and distance. For 3DMatch, we draw samples that are within 50 frames and 2 meters from the reference frame. Since the point sets in Kitti have a larger scale, we instead draw samples that are within 100 frames and 15 meters. Following~\cite{FCGF2019}, we also make sure that the samples have at least $30\%$ overlap with the reference frame. In both 3DMatch and Kitti, ground truth relative
poses are given and we use these to measure the error in rotation
angle and Euclidean distance in translation. For all methods in the evaluations, we pre-process the point sets with voxel grid downsampling. We set the voxel side length to $5$ cm for 3DMatch and $30$ cm for Kitti.

In the following sections, we first provide an ablative analysis on the 3DMatch dataset to show the impact of different components of our method. Next, we compare our approach to a number of state-of-the-art methods for pairwise and multi-view registration on both 3DMatch and Kitti. Example registrations on the Kitti dataset are visualized in Figure \ref{fig:example}. More examples and a runtime analysis are provided in the supplementary.
 
\begin{figure}
	\centering
	\renewcommand{\arraystretch}{0.2}
	\tabcolsep=0.04cm
	\begin{tabular}{ccc}
		Input & Ground truth & Ours RLL multi\\
		\includegraphics[width=0.32\columnwidth]{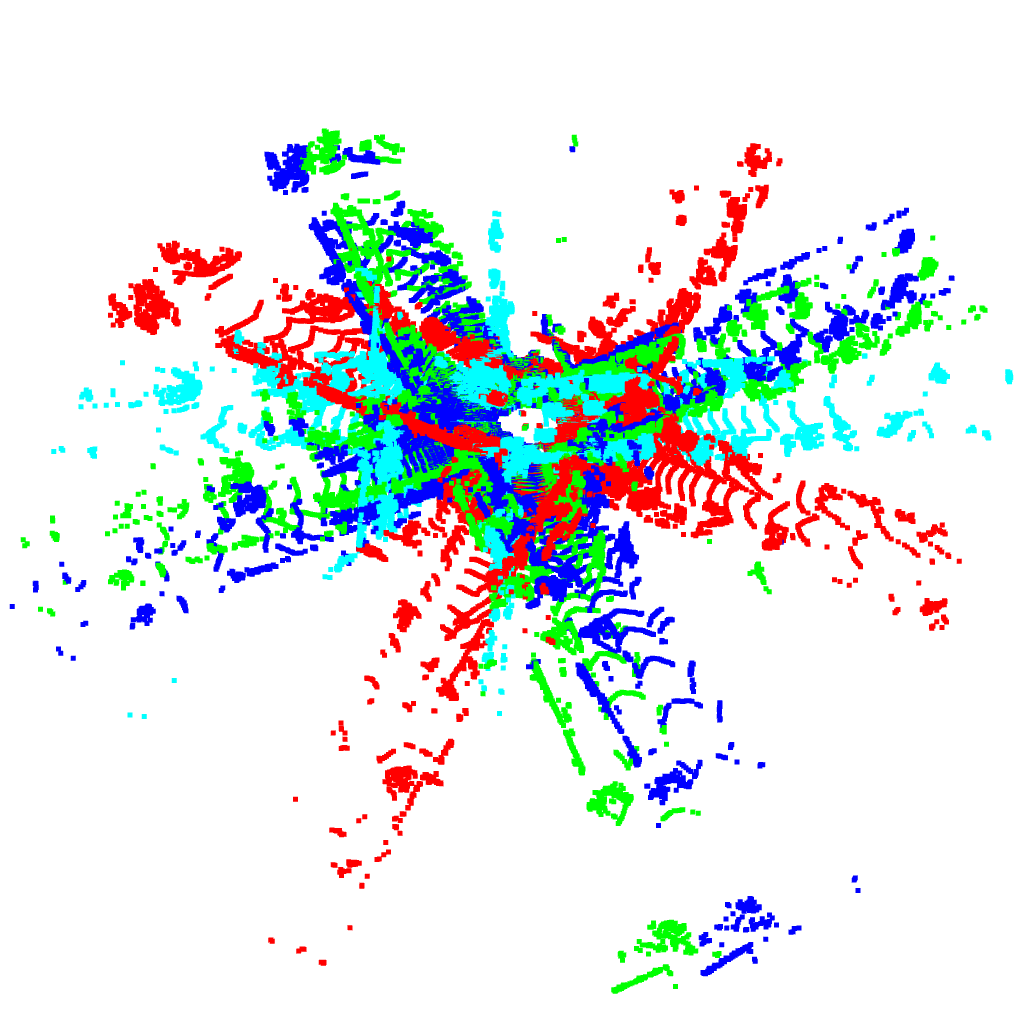}&
		\includegraphics[width=0.32\columnwidth]{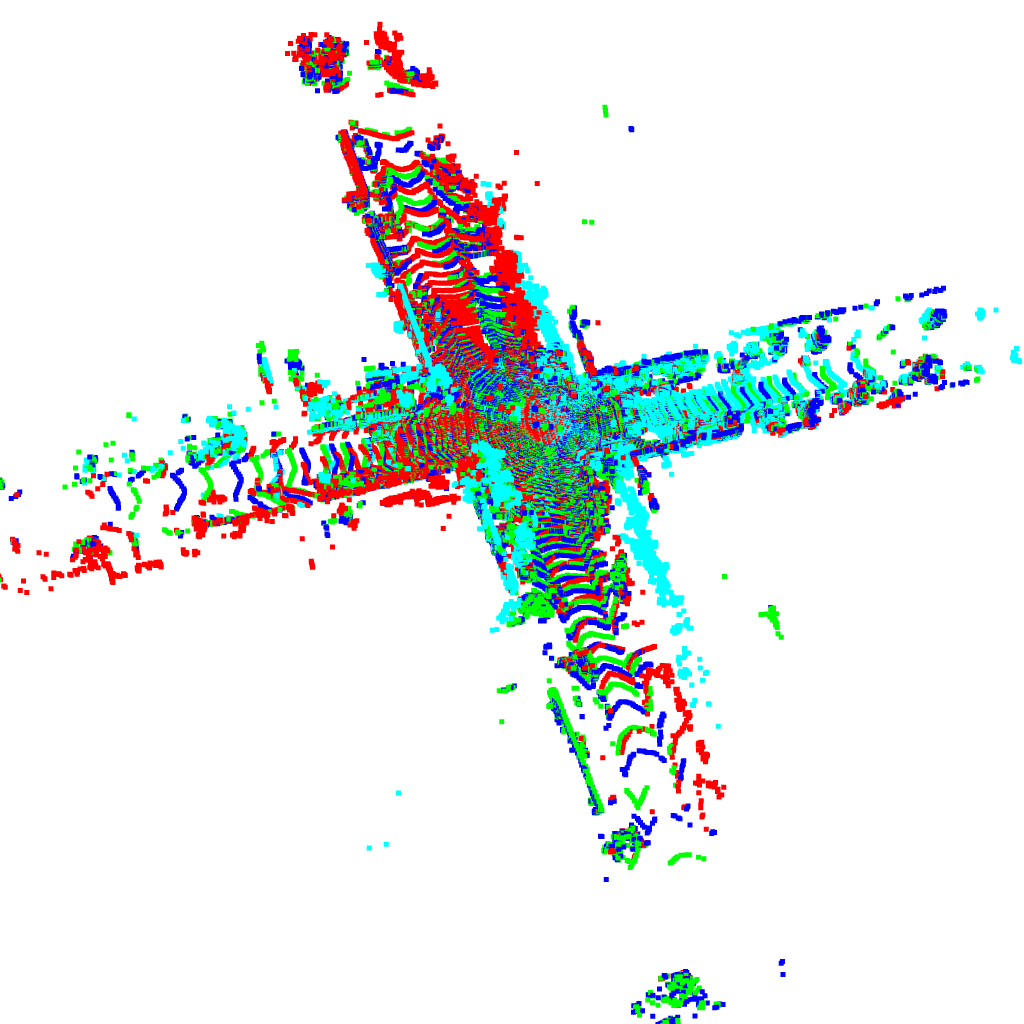}&
		\includegraphics[width=0.32\columnwidth]{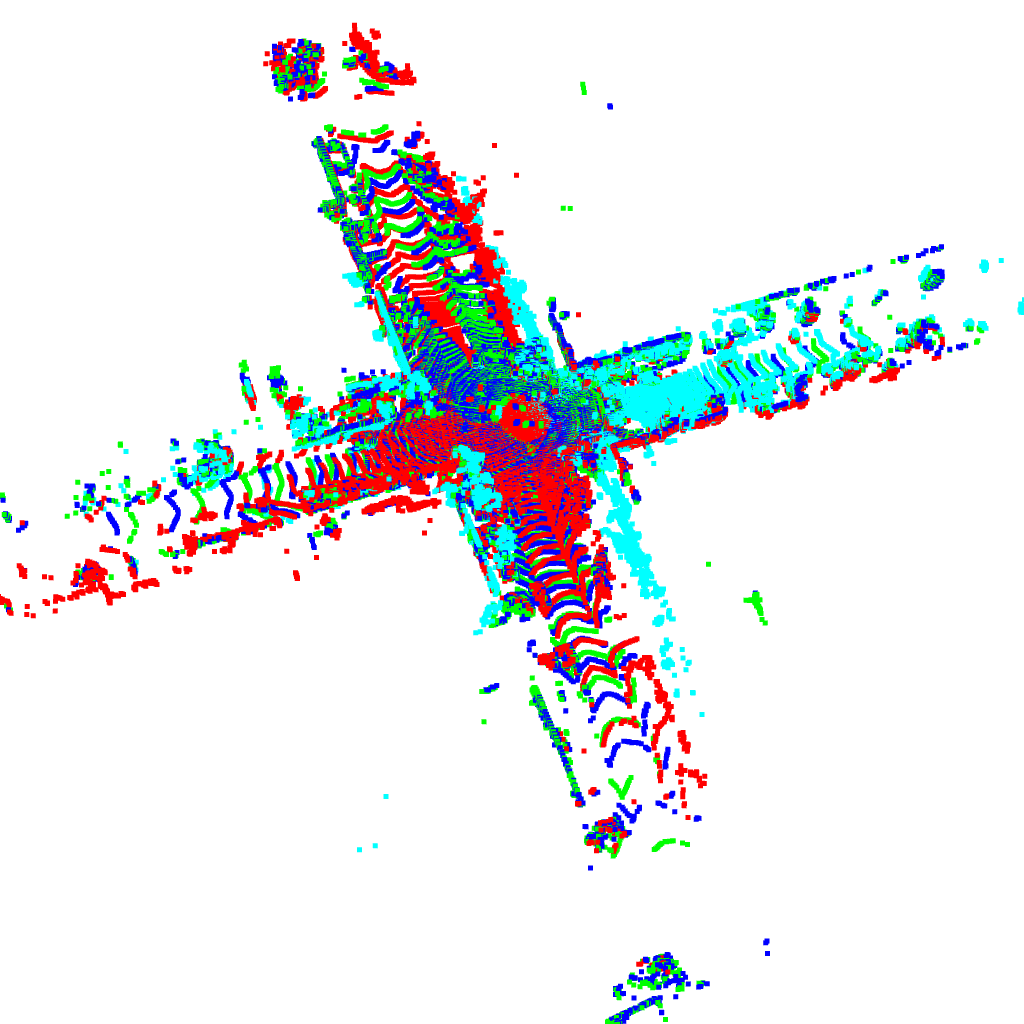}\\
		\includegraphics[width=0.32\columnwidth]{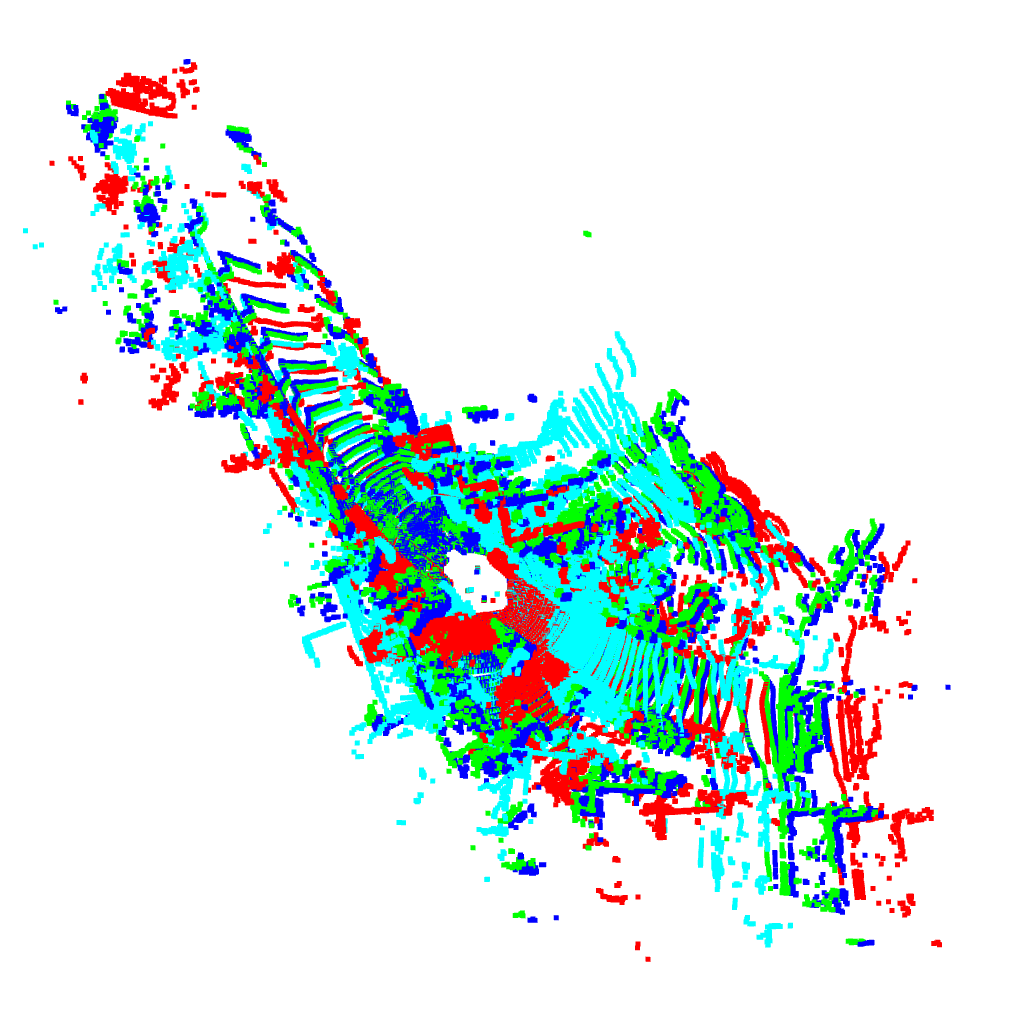}&
		\includegraphics[width=0.32\columnwidth]{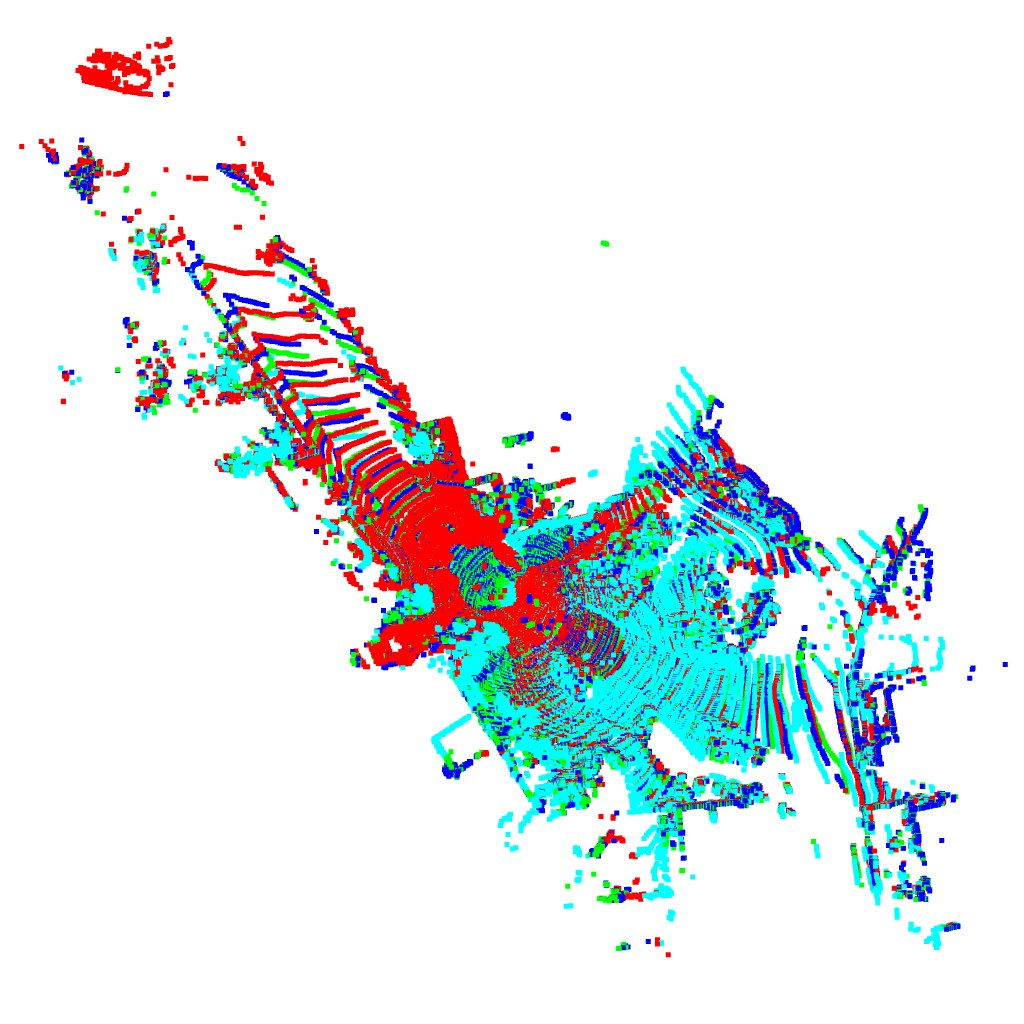}&
		\includegraphics[width=0.32\columnwidth]{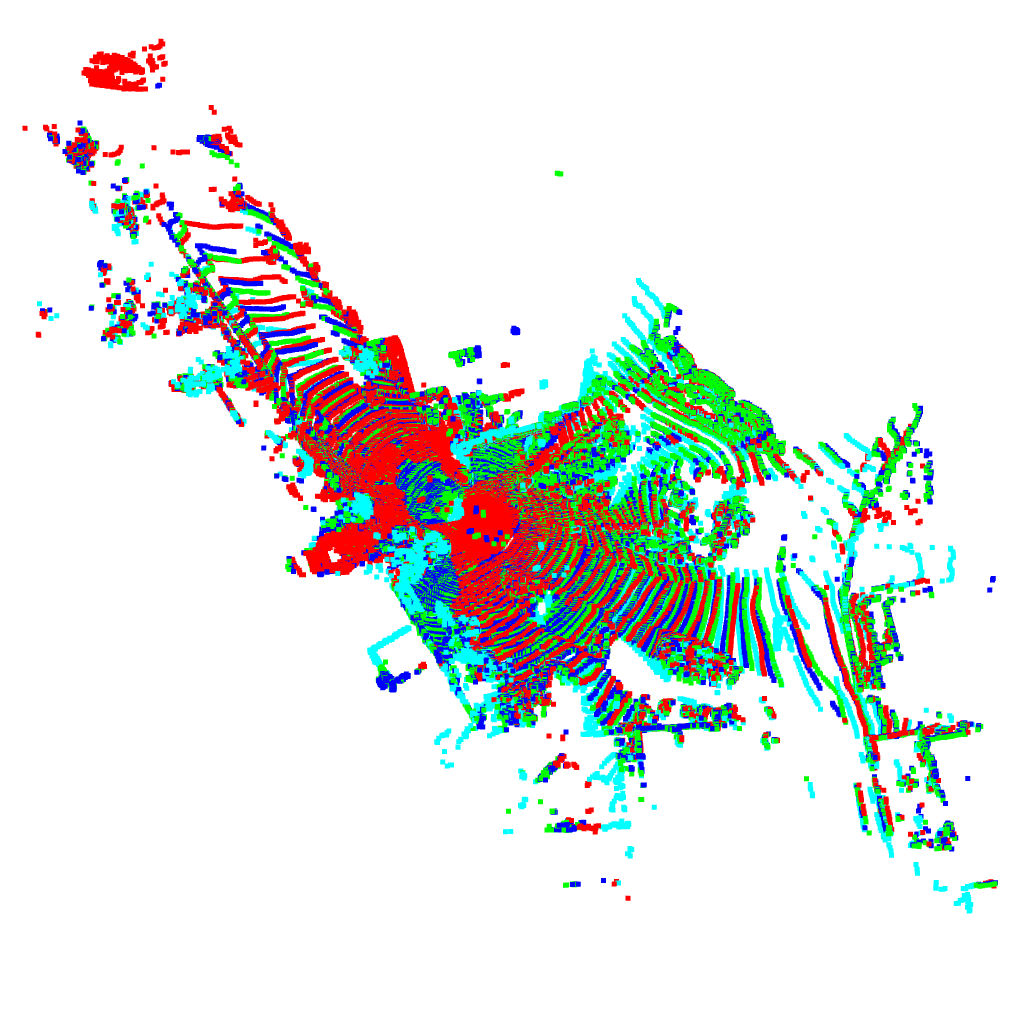}
	\end{tabular}
	\caption{Joint registration of multiple point sets from Kitti. Our approach successfully registers these point sets despite partial overlap and large deviations in initial rotations and translations.}
	\label{fig:example}
\end{figure}

\subsection{Ablative analysis}
\label{sec:ablation}

In this section we analyze the impact of the different components in our approach. First, we analyze the impact of the proposed feature model and the registration loss learning (RLL). To this end, we generate the different versions of our approach listed below.

\parsection{Baseline} As a baseline approach we employ a GMM which only models the spatial coordinates of the points. Since the downsampled RGB-D point clouds only have minor variations in density, the baseline does not benefit from density adaptive weighting~\cite{jaremo18a}. Hence, we set the weights $w$ to be equal, and this makes the baseline is equivalent to the JRMPC~\cite{evangelidis14} method. 

\parsection{Contrastive} In this version, we employ our proposed feature model (see Section. \ref{sec:feat}) using pre-trained features from~\cite{FCGF2019} with $C=32$ channels. We denote this version as ``contrastive", since the features have been obtained using contrastive learning. Like the baseline, this version does not employ any pointwise weighting.

\parsection{RLL unweighted} Next, we integrate RLL of the feature extractor (see Section \ref{sec:training}). The feature extractor is trained from scratch and is outputting $C=16$ channel feature descriptors. As in the contrastive version, we are not using any pointwise weighting here.

\parsection{RLL weighted} We further include the attention weights $w$ in the model. These are learned along with the features using RLL.

\parsection{RLL weighted 32} For fair comparison with the contrastive version, we also include a version using $C=32$ channel feature descriptors.

\parsection{RLL L2 weighted} To evaluate the impact of the robust loss $\rho$ in \eqref{eq:sample_loss}, we include a version of RLL with attention weights which have been trained using the $L^2$ loss without Geman-McClure.

We compare the above variants on the 3DMatch dataset. For simplicity, we only consider pairwise registration in this analysis. We sample 1000 pairs randomly as described above. All variants are assigned $K^{\text{run}}=100$ mixture components and we set the number of EM iterations to $N^{\text{run}}_\text{iter}=100$ during runtime. 

The ablative analysis is summarized in Table~\ref{tab:ablative}. We report the success rates, and average errors for the method variants.  A registration is considered successful if the rotation error is less than $4^\circ$ and the translation error is less than $10$ cm. The results show that the contrastive version significantly outperforms the baseline with an increased success rate of $11.9\%$. This demonstrates that the proposed feature model is able to exploit the descriptive features from the contrastive learning process. By further including the RLL, we improve the success rate over contrastive learning with $2.7\%$. The best performance is obtained when we include the learned attention weighting with an additional improvement of $2.7\%$. We further observe that increasing to $32$ channels feature descriptors do not improve the performance. Finally, we see that the proposed robust loss improves over the $L^2$ loss in success rate with $3.1\%$.  

Moreover, we analyze the impact of the number of EM iterations ($N_\text{iter}$) used during training. The above versions of our approach with RLL are trained with $N_\text{iter}=23$. In this experiment we also include versions with $9$, $15$ and $29$ iterations. From Table~\ref{tab:ablative}, we observe that all versions improve over the baseline. A significant improvement is gained by increasing $N_\text{iter}$ from $9$ to $15$. Note that, at $15$ iterations we already improve over contrastive learning. We further observe that increasing from $15$ to $29$ iterations slightly reduces the performance. The best performance, however, is achieved at $23$ iterations. 
In the following sections, registration loss learning with learned weights, and $N_\text{iter}=23$ during training, is referred to as Ours RLL.

\begin{table}
	\centering
	\sisetup{round-mode=places,round-precision=1}
	\resizebox{0.99\columnwidth}{!}{%
		\begin{tabular}{c|SSS}
			\toprule
			\text{\small method} & \text{\small  Success rate \%}& \text{\small  Avg. success err ($^\circ$)}& \text{\small  Avg. success err (cm)}\\ \midrule
			baseline&          62.463 &    1.226 &  3.241\\
			Contrastive &73.088   &  1.174  & 3.258\\
			RLL $N_\text{iter}=23$& 76.266   &  1.205 &  3.351\\
			RLL $N_\text{iter}=23$+weights& 79.543  &   1.137  & 3.079\\
			RLL $N_\text{iter}=23$+weights $C=32$ & 78.947  &   1.109  & 3.148\\
			RLL L2 $N_\text{iter}=23$+weights& 76.763  &   1.192  & 3.237\\\midrule
			RLL $N_\text{iter}=29$+weights &75.074   &  1.129  & 3.135\\
			RLL $N_\text{iter}=15$+weights &77.160   &  1.208 &  3.180\\
			RLL $N_\text{iter}=9$+weights &66.931   &  1.288 &  3.485\\
			
			\bottomrule
	\end{tabular}}
	\caption{Ablative analysis of on the 3DMatch test split. Successes are registrations with a rotation error less than four degrees and a translation error of less than 10 cm. Within the successful registrations we compute average rotation and translation errors. Our proposed feature model significantly outperforms the baseline without features. The best result is obtained by employing our proposed registration loss learning and learned attention weighting.}
	\label{tab:ablative}
\end{table}

\subsection{State-of-the-art}

\parsection{3DMatch pairwise}
We compare our approach to the state-of-the-art for pairwise registration on the 3Dmatch dataset, using the same point set pairs as in the ablative analysis. In this comparison, we include the classical ICP point-to-point and ICP point-to-plane using implementations from Open3D~\cite{Zhou2018}. We further include the probabilistic methods JRMPC~\cite{evangelidis14}, FPPSR~\cite{DanelljanICPR2016} and FilterReg~\cite{gao18filterreg} (point-to-point implementation from \cite{probreg}). For FPPSR we employ both FPFH~\cite{rusu09} and FCGF~\cite{FCGF2019} features. Additionally, we include the global correspondence based methods FGR~\cite{zhou2016fast} (Open3D) and the recent DGR~\cite{choy2020deep}, which uses deep features and a RANSAC fallback. 

\begin{figure}
	\renewcommand{\arraystretch}{0.02}
	\begin{tabular}{c}
	\includegraphics[width=\columnwidth, trim={0 2mm 0 12mm},clip]{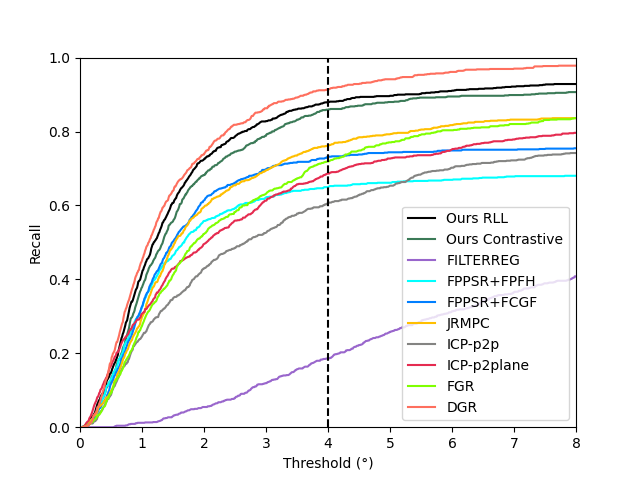}\\
	\includegraphics[width=\columnwidth, trim={0 0 0 12mm},clip]{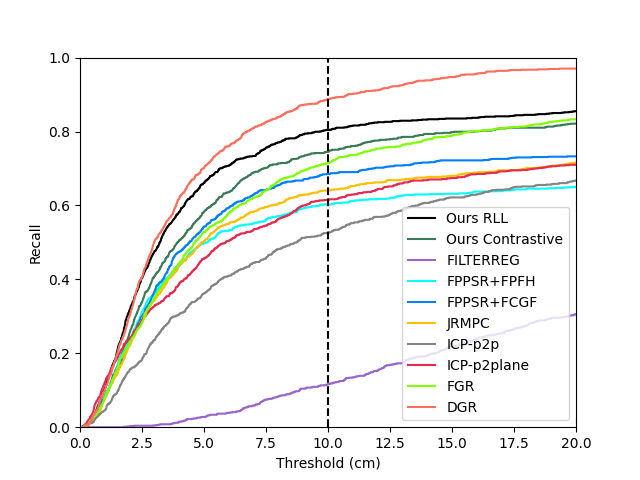}
	\end{tabular}
	\caption{Pairwise recall curves on 3DMatch for errors in rotation (top) and translation (bottom).}
	\label{fig:threedmatch-rot}
\end{figure}

The results are presented as recall curves for both the rotation and translation errors in Figure \ref{fig:threedmatch-rot}. As in the ablative analysis, we also present success rates and success mean rotation and translation errors in Table \ref{tab:threedmatch}. Additionally, we list the average runtimes over $38$ samples. For all methods, the runtimes include pre-processing, feature extraction and registration. Our approaches, DGR, FPPSR and JRMPC are run on GPU, while the other methods run on CPU. 

We observe that both the contrastive and registration loss learning versions of our approach significantly outperforms all classical and probabilistic approaches. Further, the registration loss learning version outperforms FGR with a margin of $14\%$ in success rate. Among the compared methods, our registration loss learning version is only outperformed by the recent DGR with a margin of $6\%$. On the other hand, our approach runs $2.6$ times faster than DGR, which should make it more useful in many online applications. 

\begin{table}
	\centering
	\sisetup{round-mode=places,round-precision=1}
	\resizebox{0.99\columnwidth}{!}{%
	\begin{tabular}{c|SSSS}
		\toprule
		\text{\small method} & \text{\small  Success rate \%}& \text{\small  Avg. inlier err ($^\circ$)}& \text{\small  Avg. inlier err (cm)} & \text{\small Avg. time (s)}\\ \midrule
ICP-p2p\cite{chen91}  &       48.858   &  1.325 &  3.669 & 0.550*\\
ICP-p2plane\cite{besl92} &    59.285  &   1.243  & 3.316 & 1.053*\\\midrule
FGR\cite{zhou2016fast}         &     66.137   &  1.397 &  3.431 & 0.939*\\
DGR\cite{choy2020deep}        &     85.799  &   1.155  & 3.217 & 2.549\\\midrule
JRMPC\cite{evangelidis14}       &    62.463  &   1.226 &  3.241 & 0.692\\
FPPSR+FCGF\cite{DanelljanICPR2016}  &  67.428   &  1.197  & 3.270 & 1.473\\
FPPSR+FPFH\cite{DanelljanICPR2016}  &     58.987  &   1.088 &  3.018 & 2.348\\
FILTERREG\cite{gao18filterreg}   &   7.051  &   2.029 &  6.122 & 1.346*\\\midrule
Ours Contrastive &   73.088 &    1.174  & 3.258 & 0.768\\
Ours RLL       &     79.543  &   1.137  & 3.079 & 0.773 \\
		\bottomrule
	\end{tabular}}
	\caption{Pairwise registrations on 3DMatch. Registrations with a rotation error less than four degrees and a translation error of less than 10 cm are categorized as successful. Within the successful registrations we compute average rotation and translation errors. Methods marked with * in the Avg. time column were run using CPU only, while the others were also run on GPU.}
	\label{tab:threedmatch}
\end{table}

\parsection{Kitti pairwise} 
We further evaluate our approach on the Kitti odometry test sequences for pairwise registration. Samples are generated by randomly drawing 1000 pairs as described above (see Section \ref{sec:exp}). In order to adapt our approach to the lidar scans, we perform training on the split described in Section \ref{sec:training}. We compare our approach with ICP point-to-point~\cite{chen91}, ICP point-to-plane~\cite{besl92}, DARE~\cite{jaremo18a}, FPPSR~\cite{DanelljanICPR2016}, FilterReg~\cite{gao18filterreg}, FGR~\cite{zhou2016fast} and DGR~\cite{choy2020deep}. Despite voxel grid downsampling, the density variations of the lidar point set are still high, causing the performance of the
GMM based methods to drop significantly. To counter this, we employ the density adaptive weighting in~\cite{jaremo18a} for the methods DARE, FPPSR and Ours Contrastive. We further observed that DGR struggles with finding the correct translations. Therefore, we include a version of DGR with an ICP point-to-plane refinement step, denoted DGR+ICP. 

\begin{figure}
	\renewcommand{\arraystretch}{0.02}
	\begin{tabular}{c}
	\includegraphics[width=\columnwidth,trim={0 2mm 0 12mm},clip]{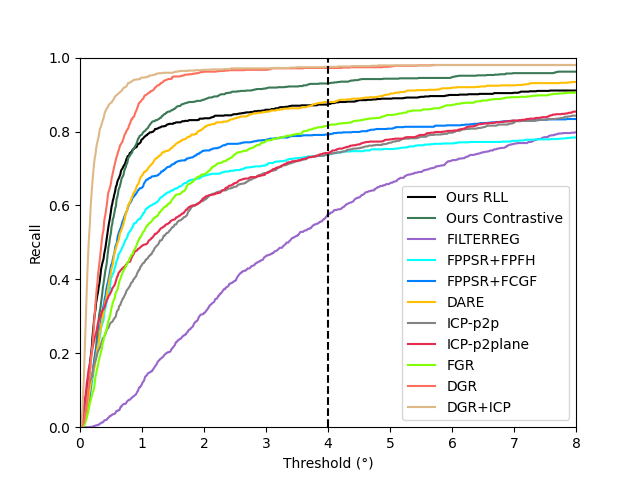}\\
	\includegraphics[width=\columnwidth, trim={0 0 0 12mm},clip]{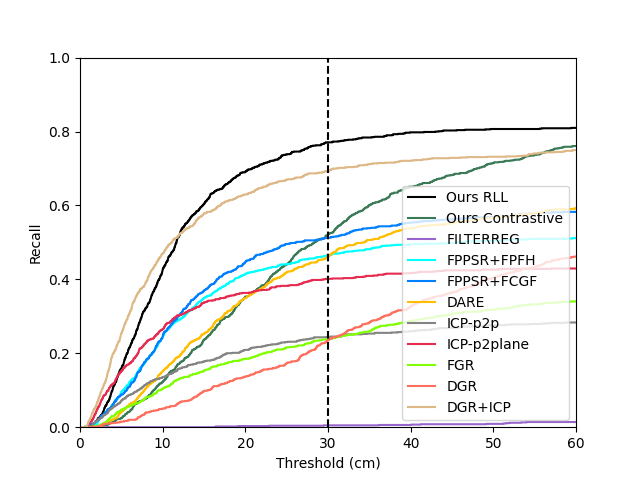}
	\end{tabular}
	\caption{Pairwise recall curves on Kitti for errors in rotation (top) and translation (bottom).}
	\label{fig:kitti-rot}
\end{figure}

Recall curves over rotation and translation errors are reported in Figure \ref{fig:kitti-rot}. In Table \ref{tab:kitti}, we report success rates, success mean rotation and translation errors\footnote{\label{note1}The Kitti dataset results reported here are slightly different from the published paper, as the experiments have been rerun. In the published version, the observation weights were missing for the FPPSR methods. Adding the weights did not change any conclusions however.}. For the Kitti results, we regard registrations with a rotation error smaller than 4 degrees and a translation error smaller than $30$ cm as being successful. We see that our approach benefits from the proposed registration loss learning, outperforming contrastive learning with a margin of $23.2\%$. Moreover, since Ours RLL is using the learned attention weighting instead of the density adaptive weighting in \cite{jaremo18a}, it also runs faster than both DARE and Ours Contrastive.

Among the previous methods DGR with ICP refinement has the highest performance. While DGR+ICP has a higher inlier rate for rotations and translation errors within $10$ cm, Ours RLL has an overall success rate which is $8.4\%$ higher. Moreover, Ours RLL runs $4.3$ times faster than DGR+ICP. 
\begin{table}
	\centering
	\sisetup{round-mode=places,round-precision=1}
	\resizebox{0.99\columnwidth}{!}{%
		\begin{tabular}{c|SSSS}
			\toprule
			\text{\small method} & \text{\small  Success rate \%}& \text{\small  Avg. inlier err ($^\circ$)}& \text{\small  Avg. inlier err (cm)} &  \text{\small Avg. time (s)}\\ \midrule
ICP-p2p\cite{chen91}     &     24.424   &  0.279 & 10.643 & 0.382*\\
ICP-p2plane\cite{besl92} &      40.140  &   0.317 &  8.818 & 0.634*\\\midrule
FGR\cite{zhou2016fast}         &     24.224  &   0.370 & 13.440 & 3.039*\\
DGR\cite{choy2020deep}         &      23.524  &   0.316  & 17.649 & 2.071\\
DGR+ICP\cite{choy2020deep}     &      69.269  &   0.136 &  8.764 & 2.466\\\midrule
DARE\cite{jaremo18a}         &    45.946   &  0.410 & 14.586 & 1.078\\
FPPSR+FCGF\cite{DanelljanICPR2016}   &    51.552  &   0.402 & 14.125& 4.492\\
FPPSR+FPFH\cite{DanelljanICPR2016}   &  38.338  &   0.390 & 14.388 & 5.811\\
FILTERREG\cite{gao18filterreg}    &    0.501  &   0.844 & 20.846 & 1.993*\\\midrule
Ours Contrastive  &    51.451 &    0.445 & 16.247& 1.217\\
Ours RLL         &     76.877   &  0.407 & 10.486&0.574\\
			\bottomrule
	\end{tabular}}
	\caption{Pairwise registrations on Kitti. Registrations with a rotation error less than four degrees and a translation error of less than 30 cm are categorized as successful. Within the successful registrations we compute average rotation and translation errors. Methods marked with * in the Avg. time column were run using CPU only, while the others were also run on GPU. }
	\label{tab:kitti}
\end{table}

\parsection{3DMatch mutli-view}
We evaluate multi-view registration on 3DMatch by creating a dataset of 500 samples with four overlapping point sets each. Each registration results in six pairwise transformations. We use these to evaluate the success rates and mean rotation and translation errors in Table \ref{tab:multi} (top). We compare our approaches to FPPSR and JRMPC since these methods directly handle multi-view registration. For this experiment we also include a version of our approach which has been trained on four views per sample, denoted Ours RLL multi. Our RLL based approaches achieve the highest success rates. We also observe that our approach benefits from training on multi-view samples.

\parsection{Kitti mutli-view}
For multi-view registration on Kitti we create a dataset of 500 samples of four overlapping point sets for each sample. As as in the 3DMatch multi-view experiment above, we use all six pairwise relative transformations in the evaluation in Table \ref{tab:multi} (bottom)\footref{note1}. We compare our approaches to FPPSR and DARE. For this experiment we also include a version of our approach that has been trained on multi-view samples, denoted Ours RLL multi. As for the pairwise experiments, these results are generated from a rerun after the paper reviews due to missing observation weigths in the FPPSR methods. Our RLL based approaches achieve the highest success rates. We further observe that Ours RLL multi outperforms Ours RLL in terms of success rate.
\begin{table}
	\centering
	\sisetup{round-mode=places,round-precision=1}
		\sisetup{round-mode=places,round-precision=1}
	\resizebox{0.99\columnwidth}{!}{%
		\begin{tabular}{c|SSSS}
			\toprule
			\multicolumn{5}{c}{\textbf{3DMatch}}\\
			\text{\small method} & \text{\small  Success rate \%}& \text{\small  Avg. inlier err ($^\circ$)}& \text{\small  Avg. inlier err (cm)} &  \text{\small Avg. time (s)}\\ \midrule
			JRMPC\cite{evangelidis14}     &          52.924  &   1.249 &  3.357  &   0.799\\
			FPPSR+FCGF\cite{DanelljanICPR2016} &        65.075  &   1.213 &  3.418  &   2.329\\
			FPPSR+FPFH\cite{DanelljanICPR2016}  &       60.104  &   1.165  & 3.240  &   4.144 \\\midrule
			Ours Contrastive  &      71.442  &   1.185 &  3.384  &   1.041\\
			Ours RLL     &      78.558   &  1.151 &  3.131  &   1.072\\
			Ours RLL multi        &     79.337 &    1.174  & 3.121  &   1.072\\
			\bottomrule
		\toprule
		\multicolumn{5}{c}{\textbf{Kitti}}\\\midrule
DARE\cite{jaremo18a}     &         37.849  &   0.414 & 14.975 &  1.366\\
FPPSR+FCGF\cite{DanelljanICPR2016} &        43.730  &   0.426 & 14.354 & 7.165\\
FPPSR+FPFH\cite{DanelljanICPR2016}  &       27.290  &   0.381 & 14.971 & 9.579 \\\midrule
Ours Contrastive  &      48.181  &   0.452  & 15.757 &  1.600\\
Ours RLL     &      68.616  &   0.411 & 11.239 &  0.807\\
Ours RLL multi        &      69.558   &  0.427 & 11.281 &  0.807\\
\bottomrule
\end{tabular}}
\caption{Multi-view registration on 3DMatch (top) and Kitti (bottom), with four point sets in each sample. For 3DMatch, registrations with a rotation error less than four degrees and a translation error of less than 10 cm are categorized as successful. Successful registrations on Kitti samples have rotation error less that four degrees and translation error less than 30 cm. Within the successful registrations we compute average rotation and translation errors.}
\label{tab:multi}
\end{table}

%% file: conclusion.tex
\section{Conclusions}
We have extended the paradigm of probabilistic point set registration to exploit the discriminative powers of learned features and weights. To learn the features and weights, we propose a registration loss learning strategy that trains the network in an end-to-end manner in a single phase. Our experiments demonstrate that the extension significantly outperforms previous probabilistic methods for both pairwise and multi-view registration.

%% file: derivations.tex
\section{Derivation of feature model updates}
In this section we derive the update step for the parameters $\nu_k$ in \eqref{eq:ml-von} in the main paper. The parameters are obtained by solving the following constrained maximization problem,

\begin{equation}
 \max_{\nu_k}  \sum_{i=1}^{M} \sum_{j=1}^{N} w_{ij} \alpha_{ijk}\nu_k^T y_{ij}\,,\;\;\text{s.t. }\|\nu_k\|=1
\label{eq:objective}
\end{equation}

Equivalently, we can set the constraint to $\nu_k ^T \nu_k = 1$. We write the resulting Lagrangian as,

\begin{equation}
L(\nu_k, \lambda)=\sum_{i=1}^{M} \sum_{j=1}^{N} w_{ij} \alpha_{ijk}\nu_k^T y_{ij} - \lambda \cdot (\nu_k ^T \nu_k -1)\,,
\label{eq:lagrangian}
\end{equation}
where $\lambda \in \reals$ is the Lagrange multiplier. At solutions to
\eqref{eq:objective}, the gradient of \eqref{eq:lagrangian} should vanish, i.e.~ 
\begin{equation}
\nabla_{\nu_k} L(\nu_k,\lambda)=\sum_{j=1}^{N} w_{ij} \alpha_{ijk}y_{ij}-2\lambda \nu_k = 0\,.
\end{equation}
We solve this equation for $\nu_k$ as,

\begin{equation}
\nu_k = \frac{\sum_{j=1}^{N} w_{ij} \alpha_{ijk}y_{ij}}{2\lambda}\,.
\end{equation}

We see that the constraint is satisfied when $\lambda=\big||\sum_{j=1}^{N} w_{ij} \alpha_{ijk}y_{ij} \big||/2$ and we arrive at the expression in \eqref{eq:ml-von}.

%% file: runtime.tex
\section{Runtime analysis}
We here provide an empirical runtime analysis of our method for pairwise registration on 3DMatch point sets. The computed values are averages over $95$ samples. The main processing steps in our method are the initial voxel downsampling, the feature extraction and the registration step. In our implementation, the voxel downsample step accounts for $45.7\%$ of the total runtime. The subsequent feature extraction steps takes $9.6\%$ of the runtime. Most of the remaining time is due to the registration step at $44.2\%$. Note that this fraction can be made smaller by employing fewer EM iterations. Another small fraction is caused by overhead operations and memory management.

Timings in seconds are provided in the main paper (Tables \ref{tab:threedmatch}, and \ref{tab:kitti}).

%% file: qualitative.tex
\section{Qualitative examples}
\begin{figure}
	\begin{tabular}{cc}
	\includegraphics[width=0.49\columnwidth]{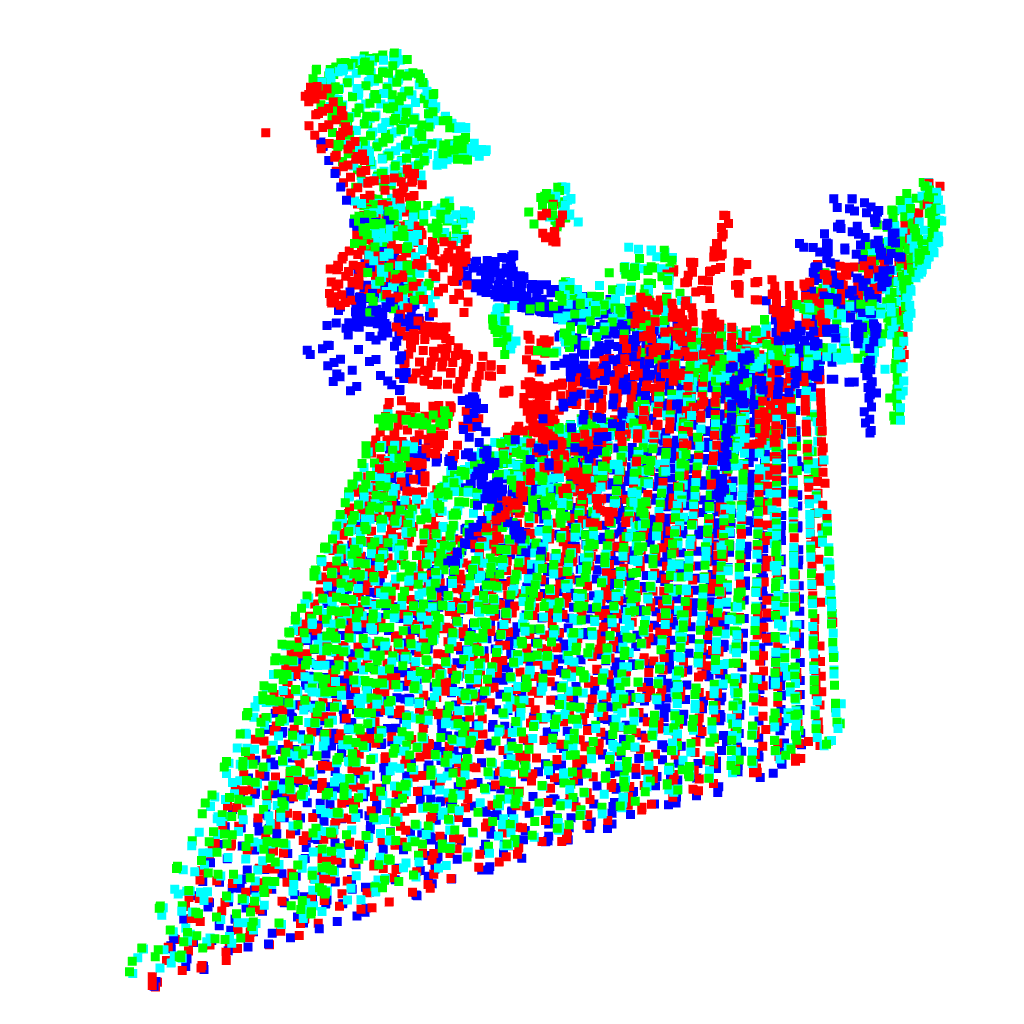}&
	\includegraphics[width=0.49\columnwidth]{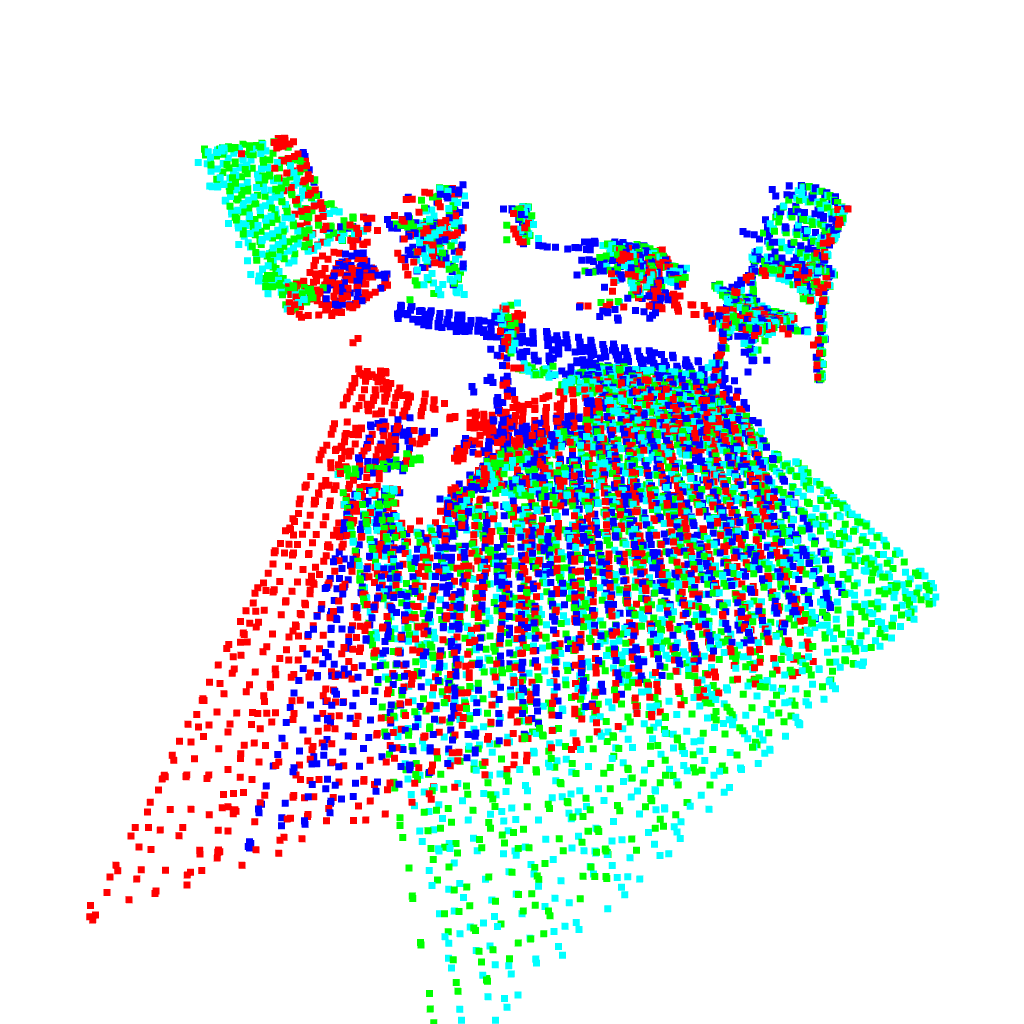}\\
	Input&Ground-truth\\
	\includegraphics[width=0.49\columnwidth]{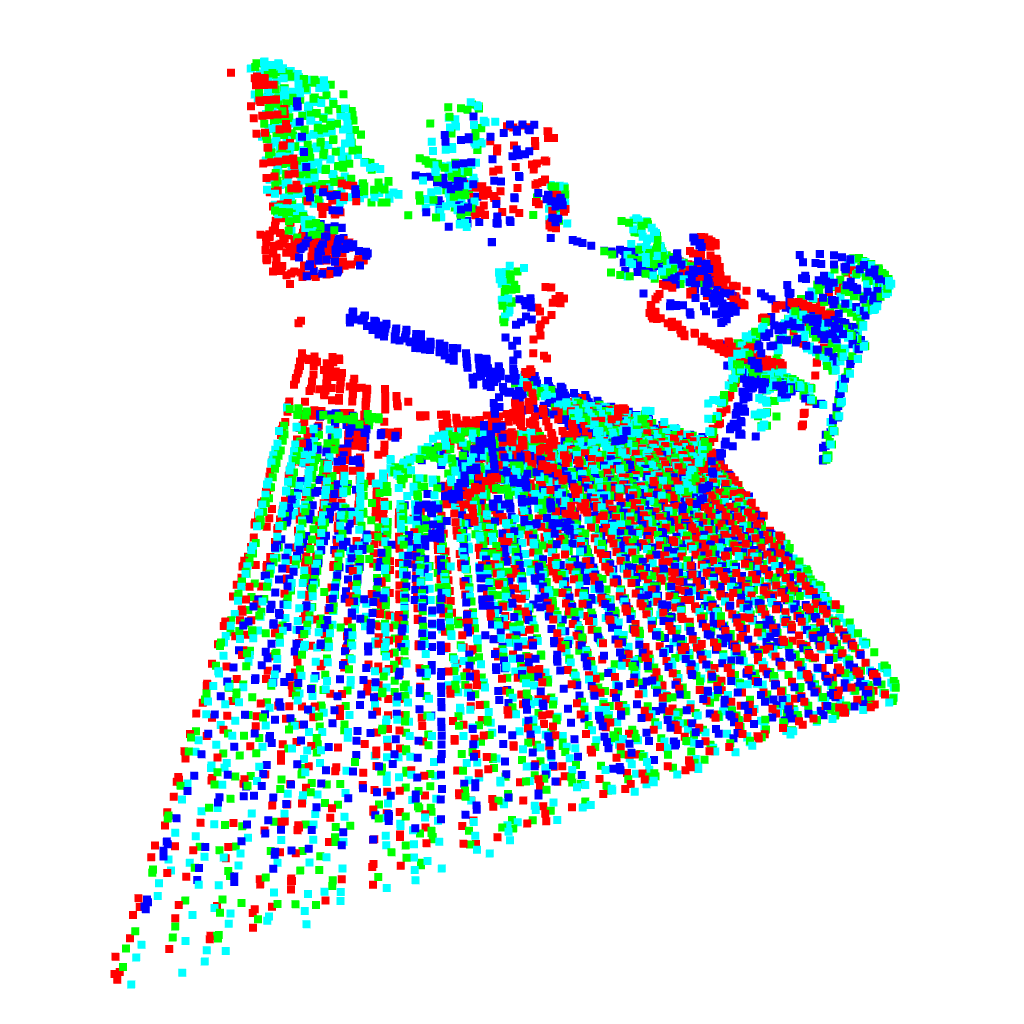}&
	\includegraphics[width=0.49\columnwidth]{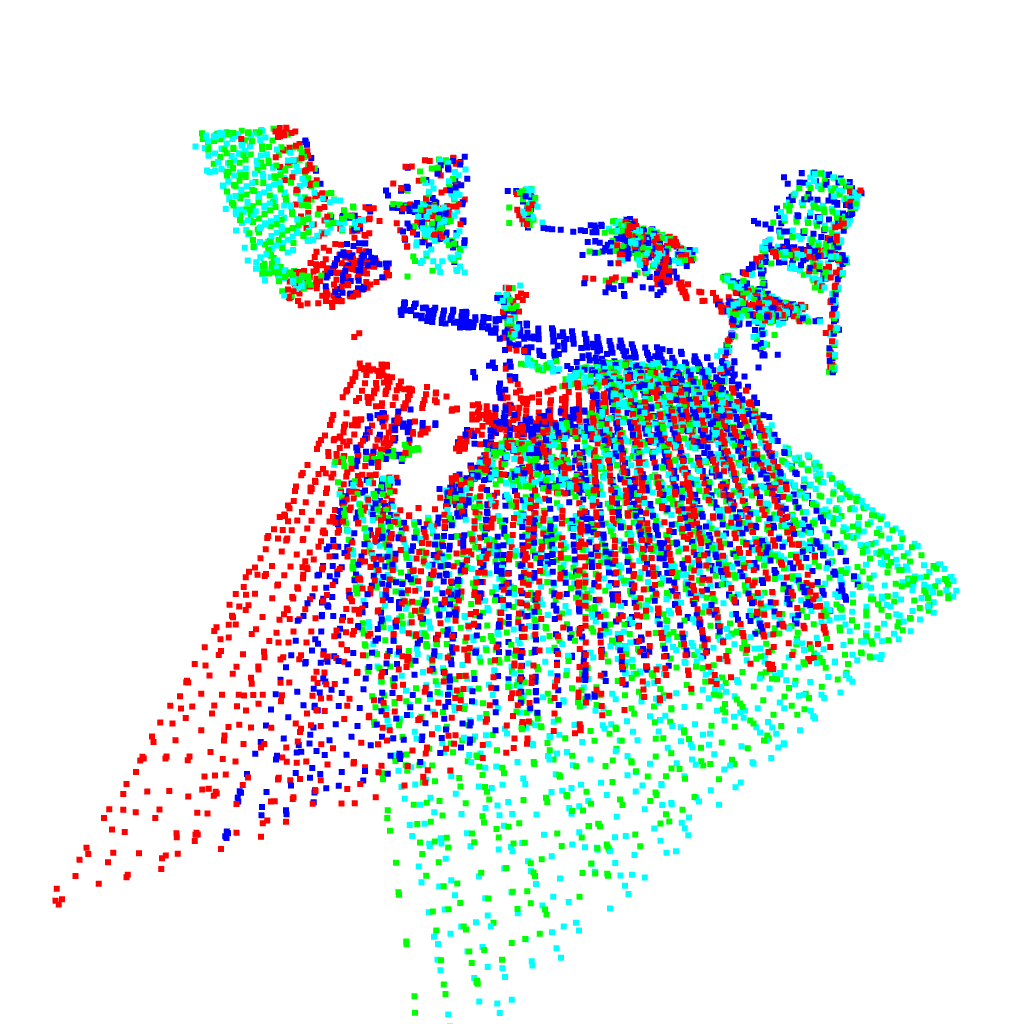}\\
	JRMPC&Ours RLL Multi
	\end{tabular}
\caption{Example of joint registration of four point sets from 3Dmatch.}
\label{fig:3dmatch}
\end{figure}

\begin{figure}
	\begin{tabular}{cc}
		\includegraphics[width=0.49\columnwidth]{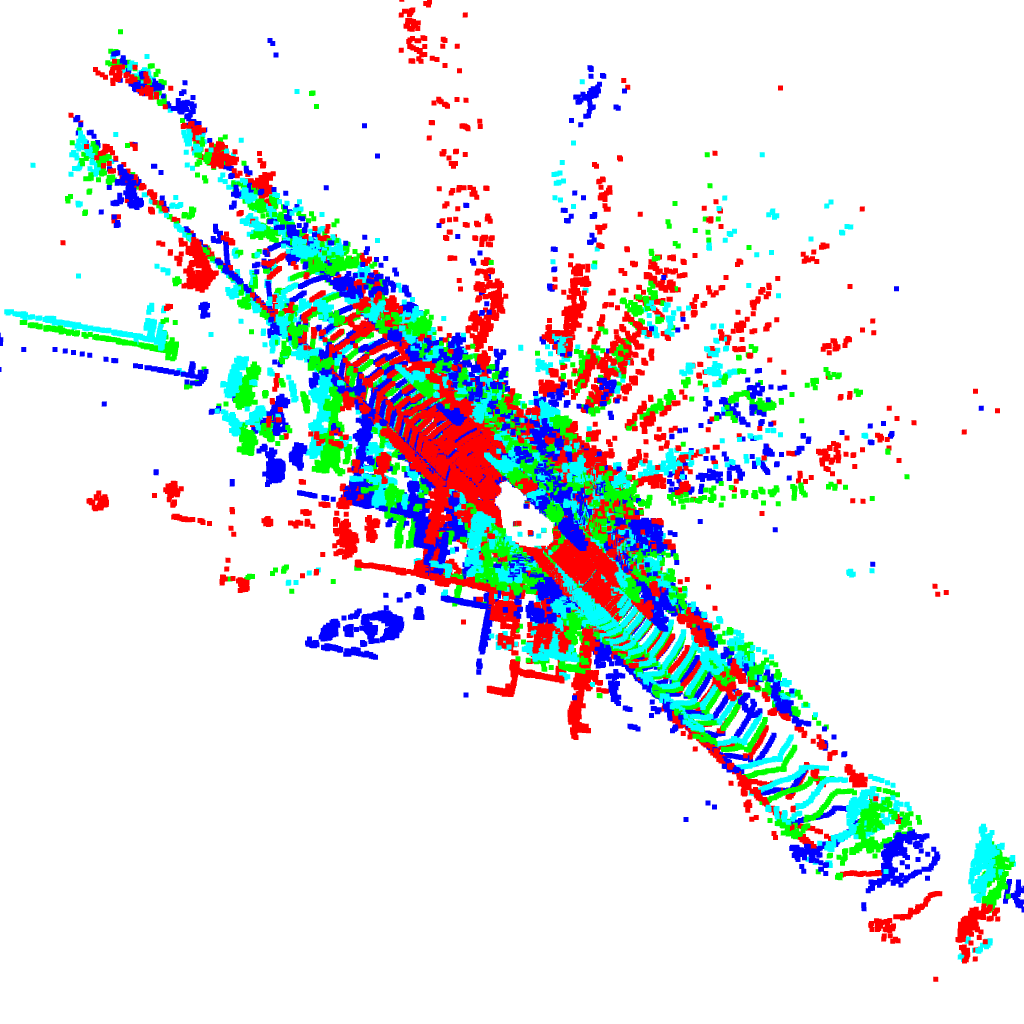}&
		\includegraphics[width=0.49\columnwidth]{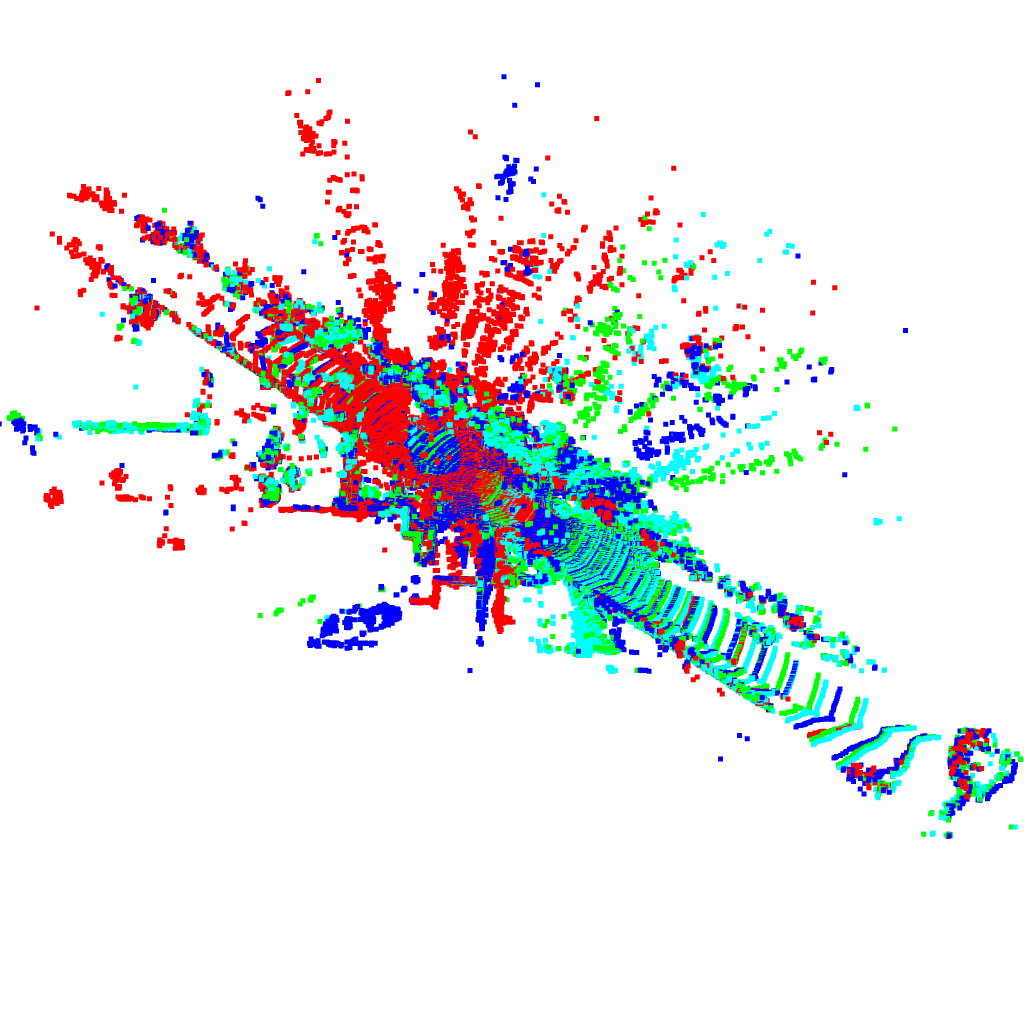}\\
		Input&Ground-truth\\
		\includegraphics[width=0.49\columnwidth]{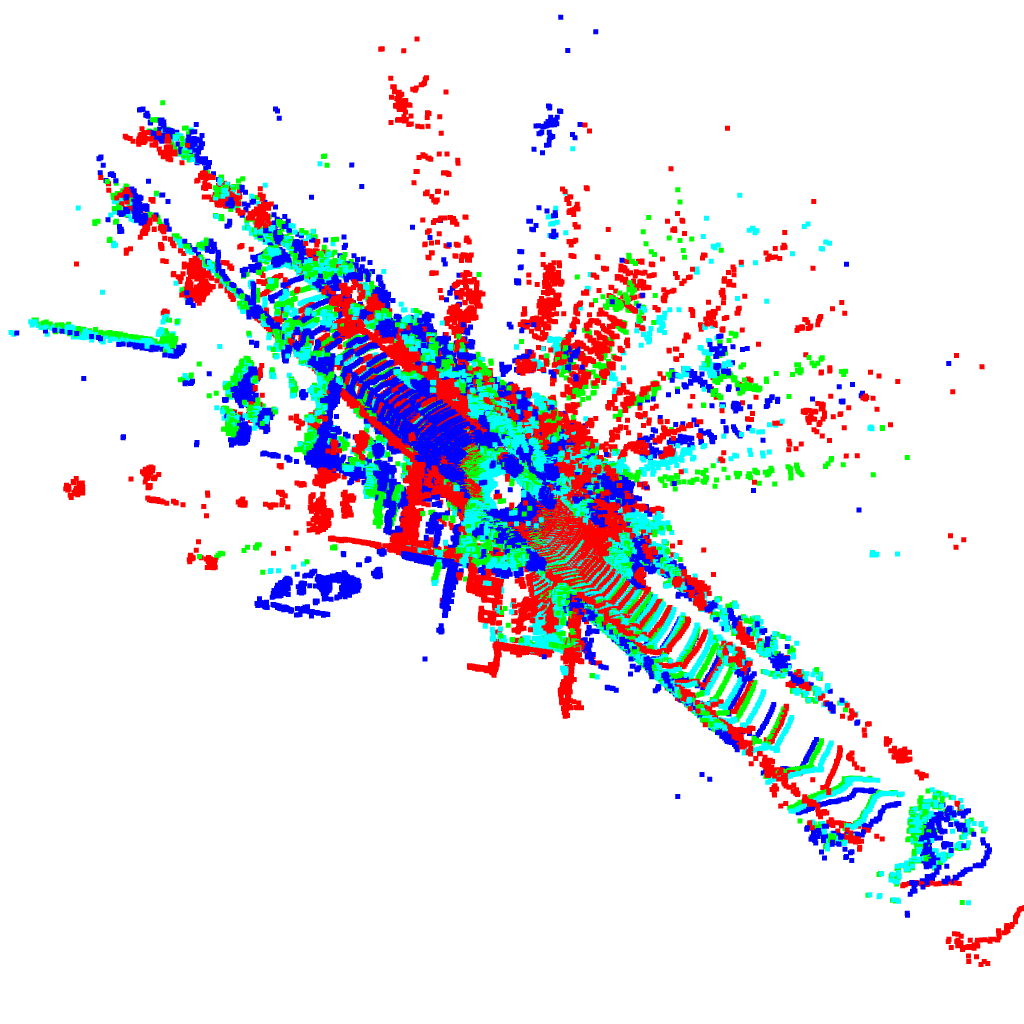}&
		\includegraphics[width=0.49\columnwidth]{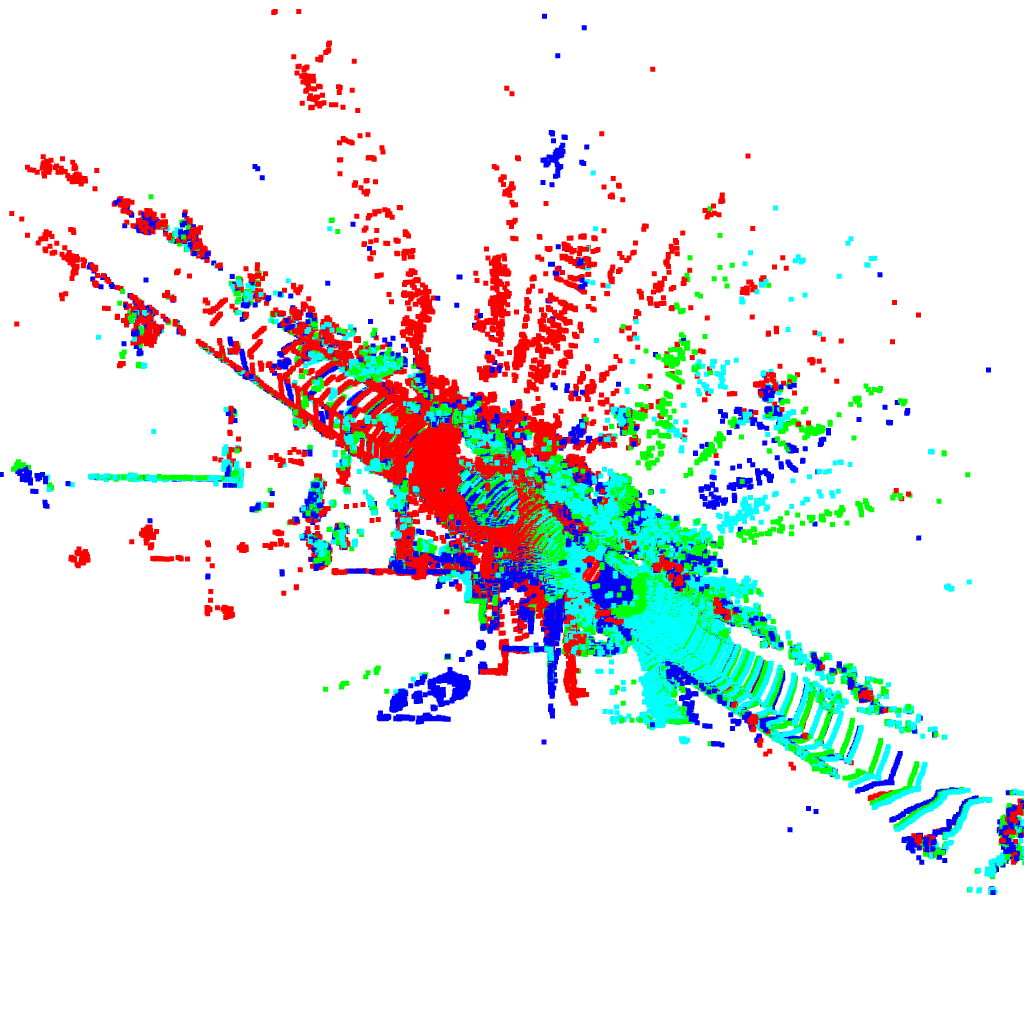}\\
		DARE&Ours RLL Multi
	\end{tabular}
	\caption{Example of joint registration of four point sets from Kitti.}
	\label{fig:kitti}
\end{figure}

We visualize examples of successful registrations produced by our
approach on multi-view point sets from 3DMatch (Figure \ref{fig:3dmatch}) and Kitti (Figure \ref{fig:kitti}). We see that the baseline approaches JRMPC~\cite{evangelidis14} and DARE~\cite{jaremo18a}, which are not using the proposed feature model, fail to register these point sets. 

\begin{figure}
	\includegraphics[width=0.49\columnwidth]{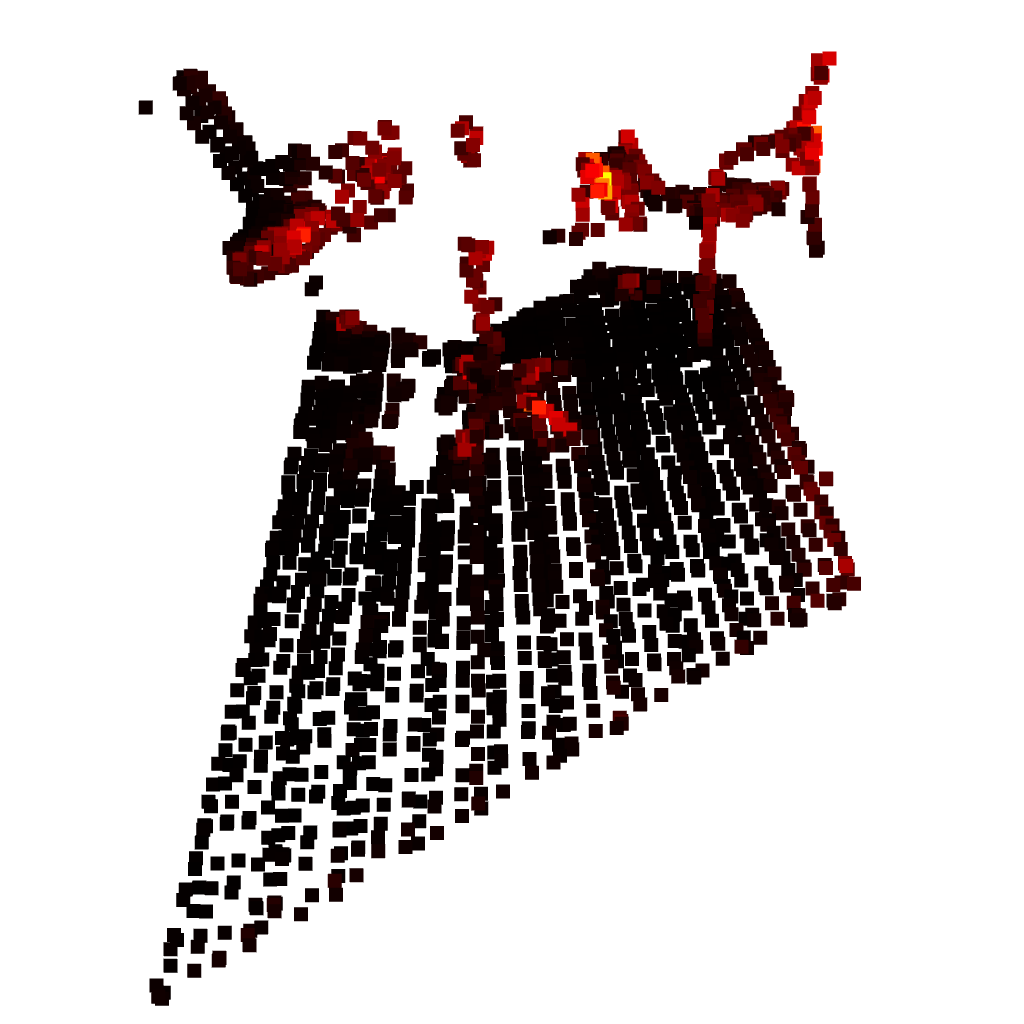}
	\includegraphics[width=0.49\columnwidth]{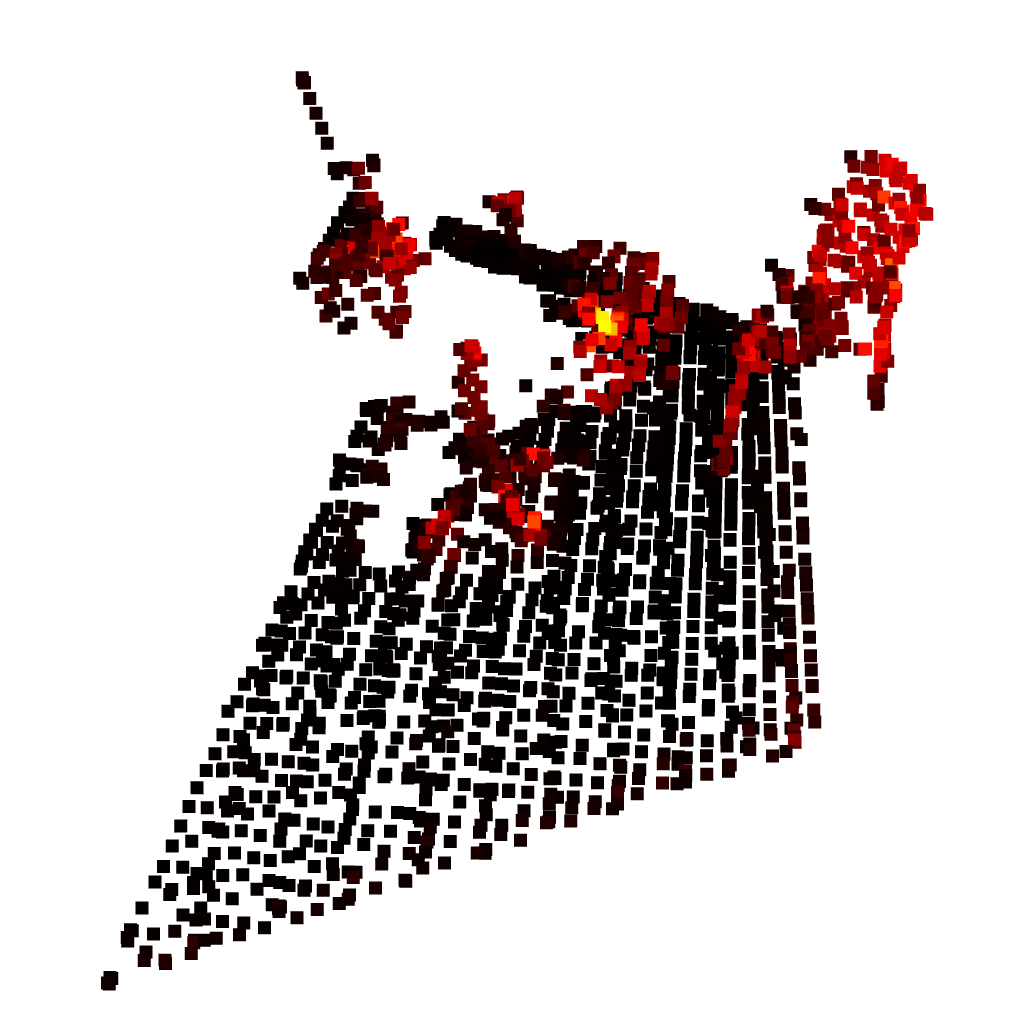}
	\includegraphics[width=0.49\columnwidth]{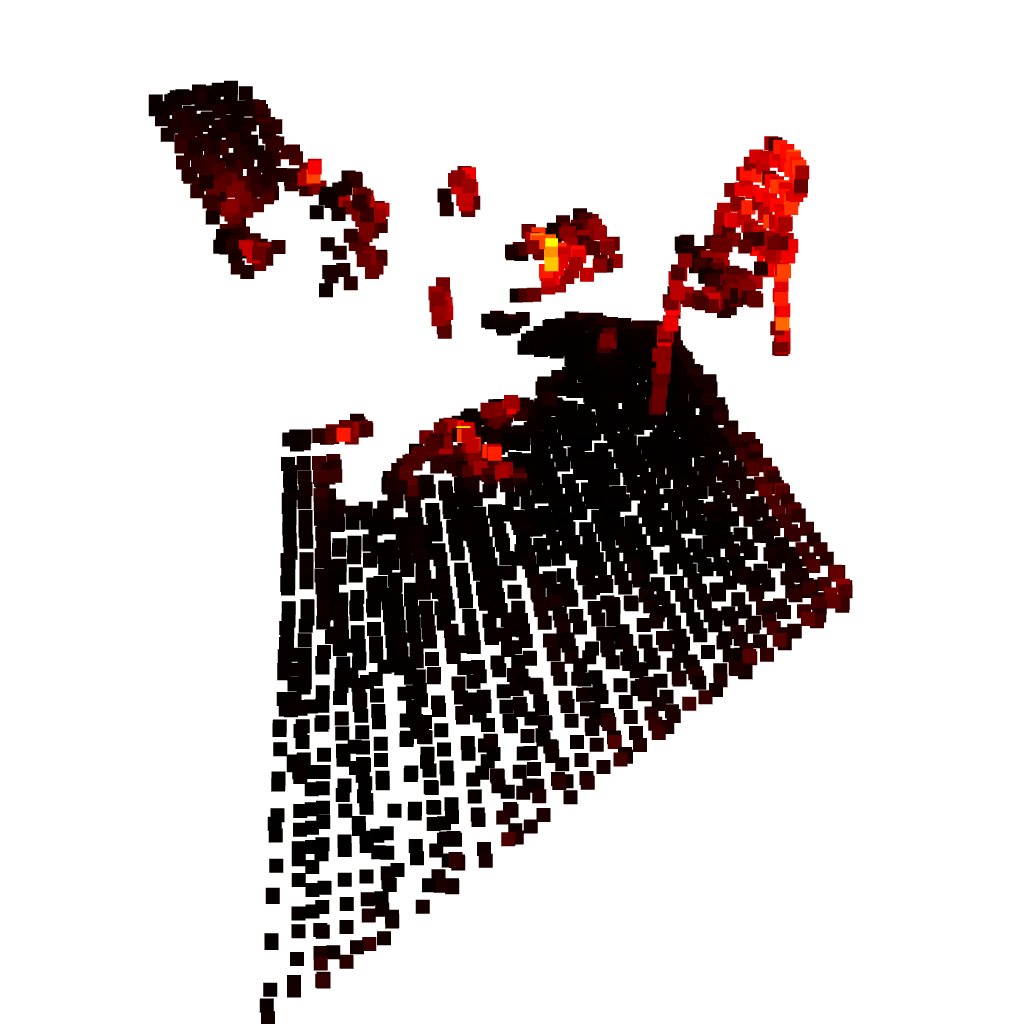}
	\includegraphics[width=0.49\columnwidth]{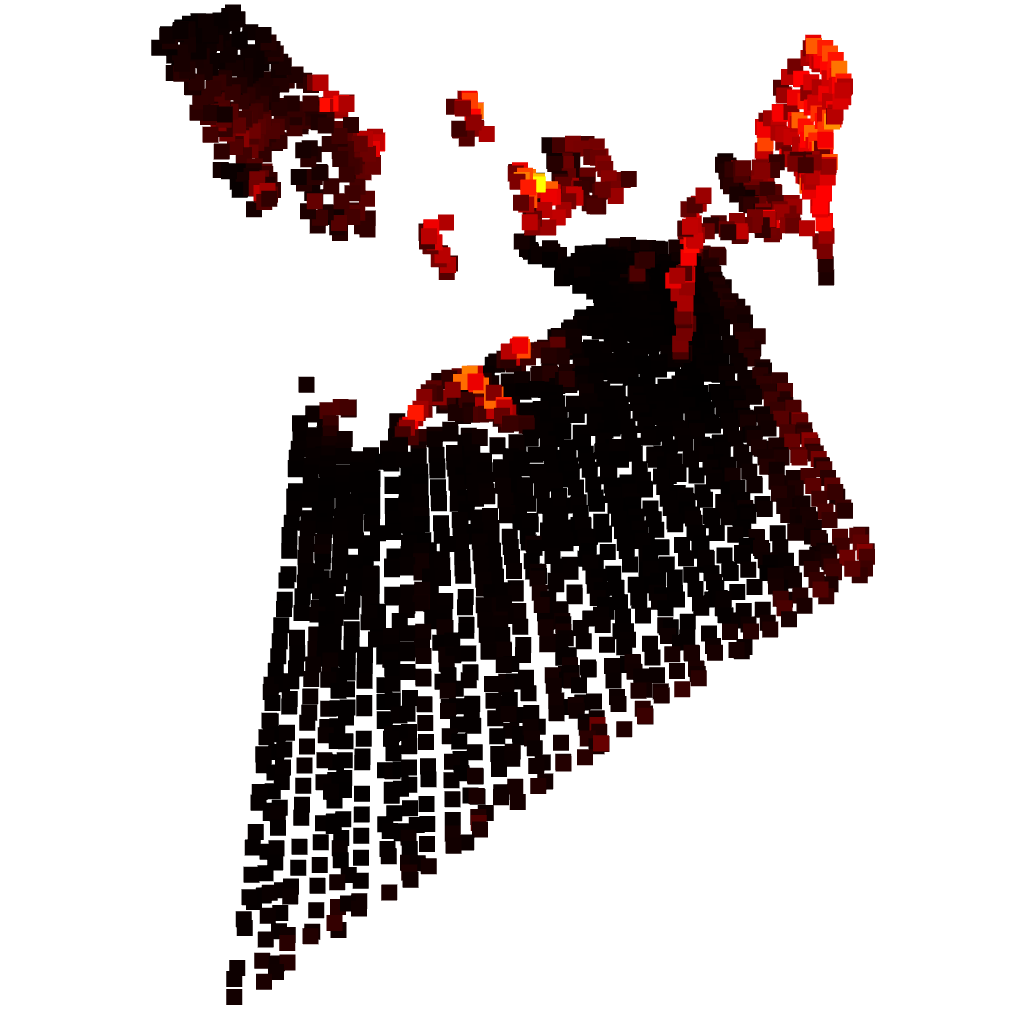}
	\caption{Visualization of the learned point-wise weighting. Brighter colors indicate higher weights.}
	\label{fig:3dmatch-w}
\end{figure}

\begin{figure}
	\includegraphics[width=0.49\columnwidth]{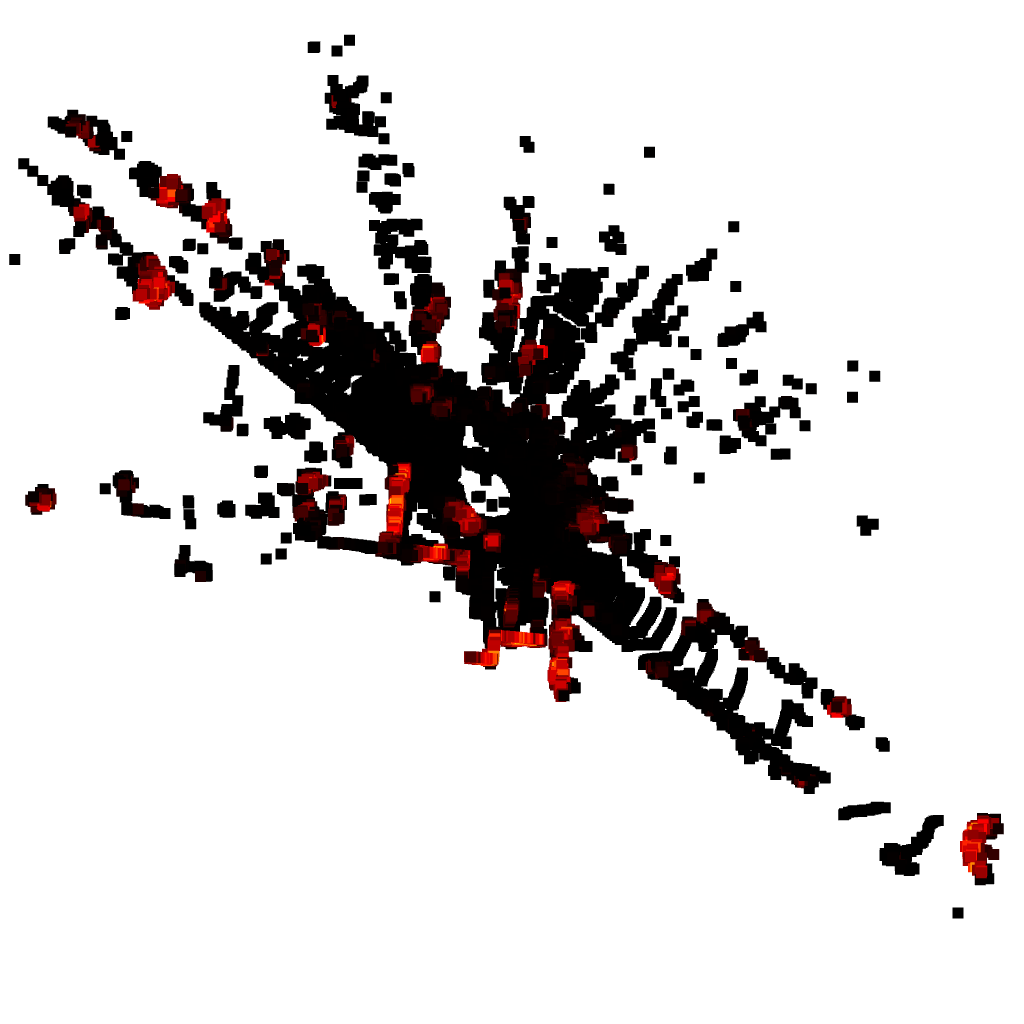}
	\includegraphics[width=0.49\columnwidth]{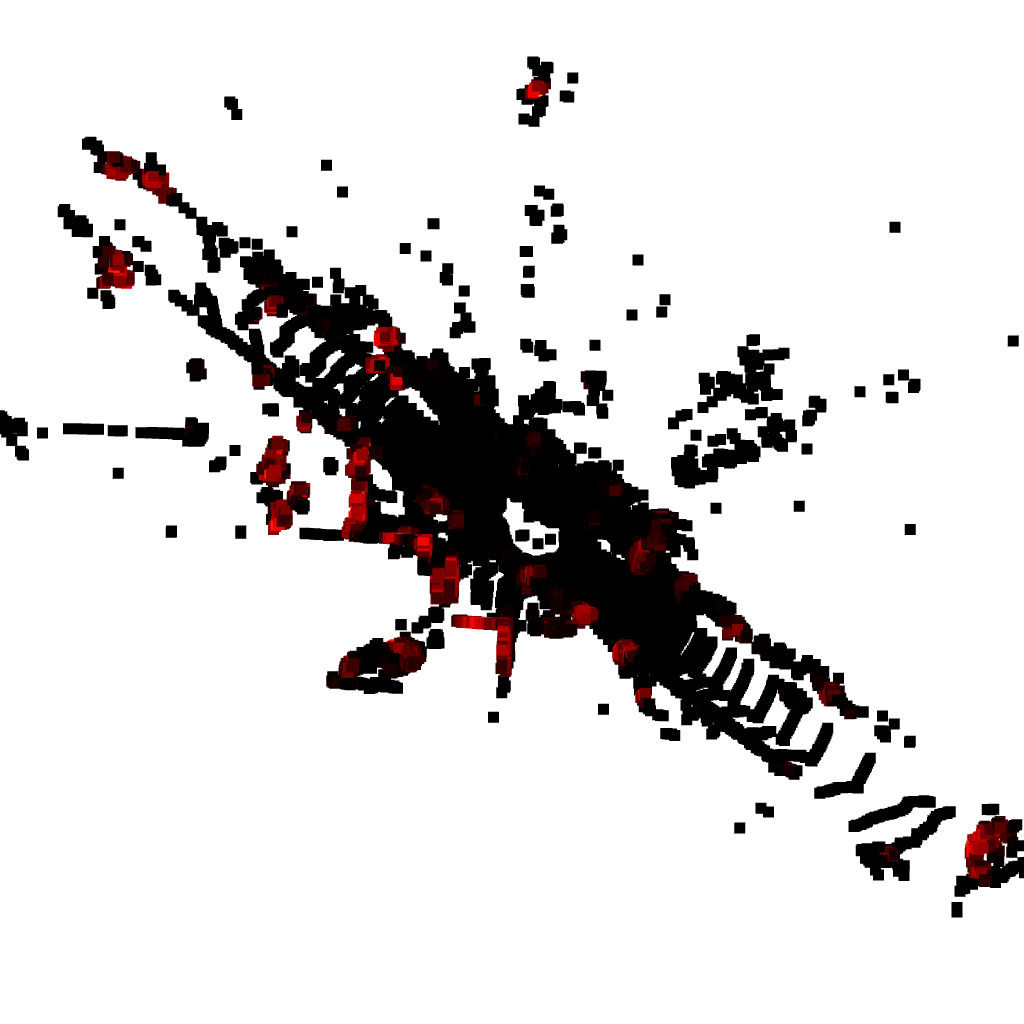}
	\includegraphics[width=0.49\columnwidth]{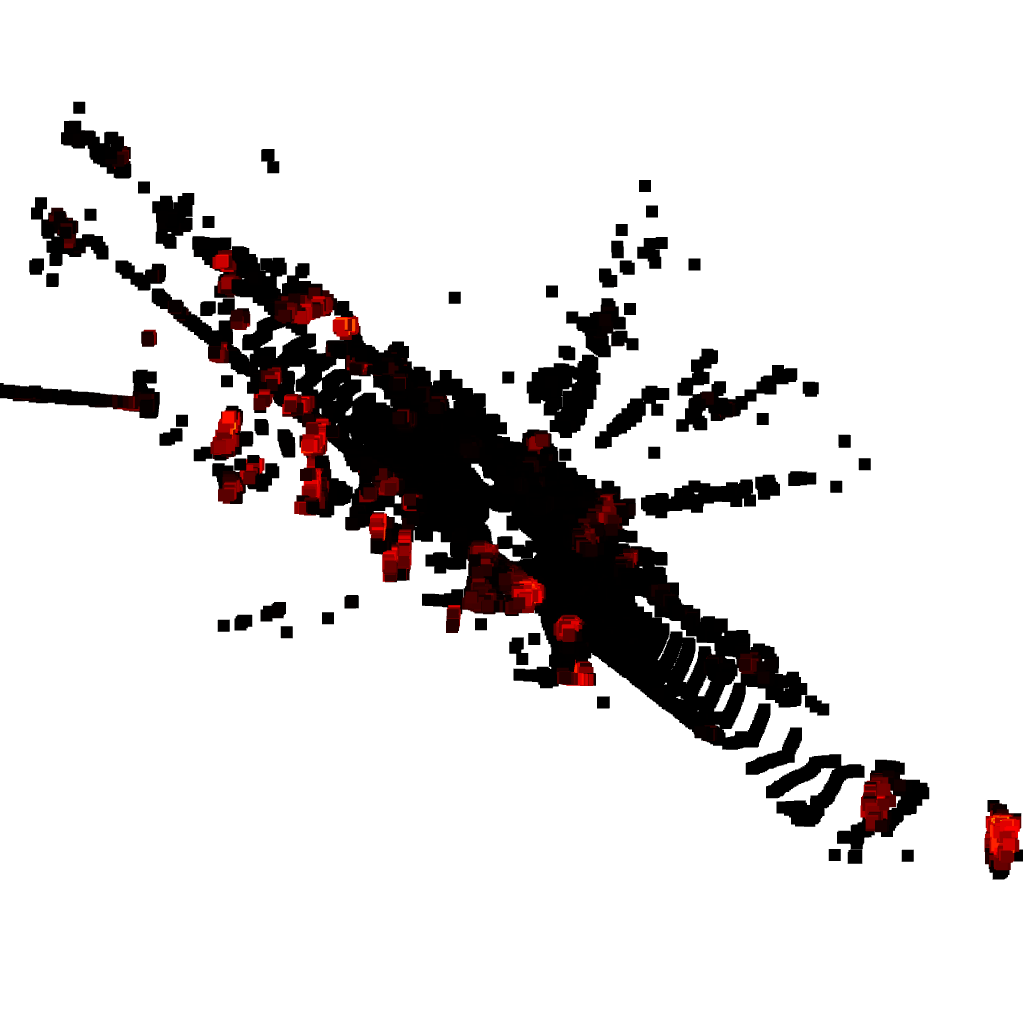}
	\includegraphics[width=0.49\columnwidth]{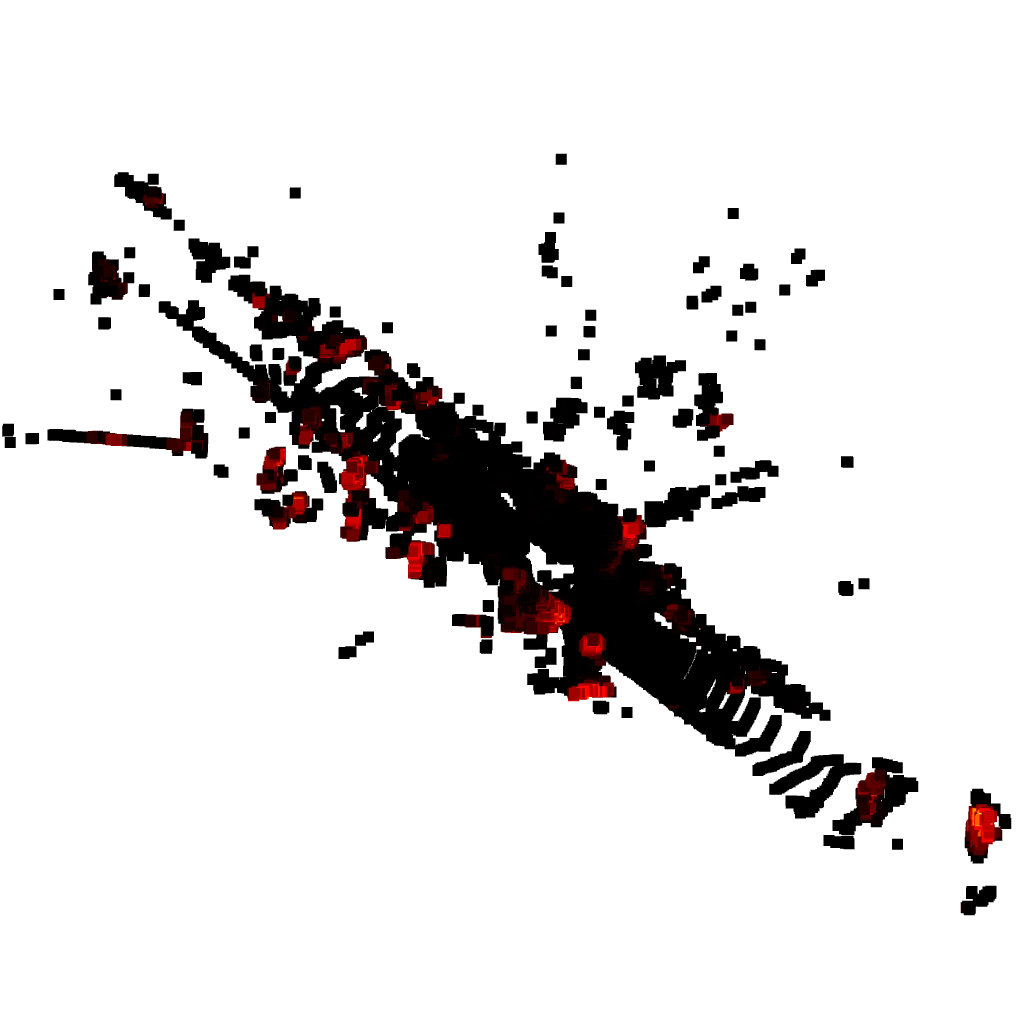}
	\caption{Visualization of the learned point-wise weighting. Brighter colors indicate higher weights.}
	\label{fig:kitti-w}
\end{figure}

We also visualize the corresponding predicted weights in Figures \ref{fig:3dmatch-w} and \ref{fig:kitti-w}. In both cases, the weighting focuses on specific structures and objects, while reducing the impact of ambiguous flat regions such as floors and ground.

%% file: experiments_supp.tex
\section{Experiments}
\begin{figure}[t!]
	\centering
	\renewcommand{\arraystretch}{0.02}
	\begin{tabular}{@{}c@{}c@{}}
		\includegraphics[width=0.49\columnwidth,trim={7mm 0 12mm 12mm},clip]{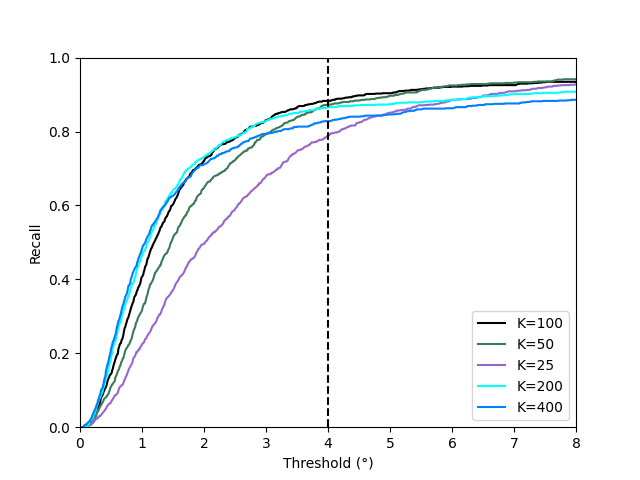}&
		\includegraphics[width=0.49\columnwidth, trim={7mm 0 12mm 12mm},clip]{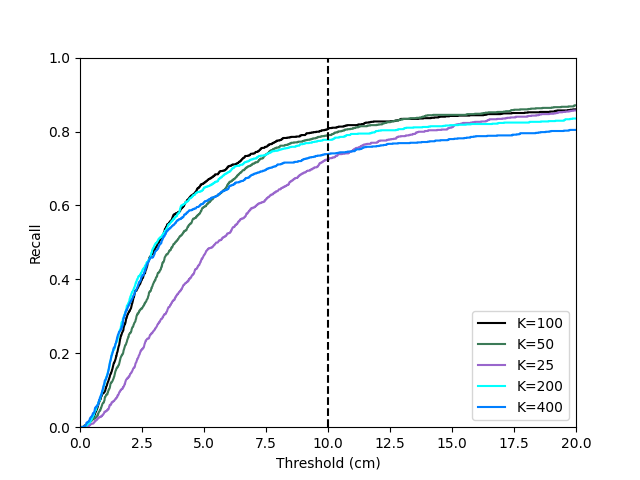}\\
	\end{tabular}
	\caption{Pairwise recall curves on 3DMatch for errors in rotation (left) and translation (right). The curves show in recall for different number of mixture components $K$. }
	\label{fig:num-comp}
\end{figure}

In this section we provide additional results that did not fit into the experiments section of the main paper. To analyze of the impact of the number of mixture components $K$, we compare the performance of our method (RLL $N_iter = 23+$weights) with $K=25,50,100,200,400$ on the 3D Match test dataset. The results are presented in Figure~\ref{fig:num-comp} as recall curves. As we can see, using $K=400$ components gives the highest recall at smaller error thresholds for both rotation and translation. However, at larger errors using $K=100$ components gives higher recall.

\begin{table}
	\centering
	\resizebox{0.99\columnwidth}{!}{%
		\begin{tabular}{c|cccccc}
			\toprule
			\text{\small method \textbackslash{} 3DMatch} & \text{\small  rot. inlier rate \%}& \text{\small  min rot. inlier rate \%}& \text{\small  max rot. inlier rate \%} &  \text{t. inlier rate \%} & \text{min t. inlier rate \%} & \text{max t. inlier rate \%}\\ \midrule
			JRMPC\cite{evangelidis14}     &         65.9 &        96.9 &        45.2  &    99.6    &     98.6   &      50.5\\
			FPPSR+FCGF\cite{DanelljanICPR2016} &        71.8      &   98.4    &     52.8   &   95.1    &     99.6     &    55.9\\
			FPPSR+FPFH\cite{DanelljanICPR2016}  &      66.3   &      97.5   &      46.2    & 94.6     &    99.0    &     47.6 \\\midrule
			Ours Contrastive  &      80.8   &    99.0   &      65.5   &   99.8    &     100.0   &      71.9\\
			Ours RLL     &      87.5   &      99.6   &      75.4   &   99.9    &     100.0   &      77.6\\
			Ours RLL multi        &     87.2   &      99.2       &  75.2   &   99.8   &      100.0   &      78.4\\
			\bottomrule
	\end{tabular}}
	\resizebox{0.99\columnwidth}{!}{%
		\begin{tabular}{c|cccccc}
			\toprule
			\text{\small method \textbackslash{} Kitti} & \text{\small  rot. inlier rate \%}& \text{\small  min rot. inlier rate \%}& \text{\small  max rot. inlier rate \%} &  \text{t. inlier rate \%} & \text{min t. inlier rate \%} & \text{max t. inlier rate \%}\\ \midrule
			DARE\cite{jaremo18a}     &        81.6    &     98.8     &    67.1    &  57.9    &     85.4       &  11.1\\
			FPPSR+FCGF\cite{DanelljanICPR2016} &        77.8    &     97.7   &      61.8   &   64.2    &     88.3    &     15.6\\
			FPPSR+FPFH\cite{DanelljanICPR2016} &      64.3   &      95.3    &     41.9  &    49.5     &    72.7    &     7.4 \\\midrule
			Ours Contrastive  &      0.901    &     100.0     &    81.7   &   76.8     &    88.7     &    16.8\\
			Ours RLL     &     82.5   &      99.6   &      70.2  &    78.1    &     97.9    &     43.7\\
			Ours RLL multi        &     83.5   &      99.4     &    71.2     & 78.7      &   98.4      &   43.7\\
			\bottomrule
	\end{tabular}}
	\caption{Multi-view registration with four point sets in each
          sample for 3DMatch (top) and Kitti (bottom). Inlier/outlier
          splits are as in the main paper. The min and max columns
          only consider the min and max error pair in each sample respectively.}
	\label{tab:maxmin-multi}
\end{table}

We also extend the results provided for multi-view registration in Table \ref{tab:multi} in the paper. In Table~\ref{tab:maxmin-multi}, we include inlier rates for both rotation and translations\footref{note1}. As in the paper, we use thresholds of 4 degrees for the rotations errors and 10 cm for translation on 3D Match and 30 cm for translation on Kitti. In addition we also show maximum and minimum inlier rates, which are the inlier rates for the maximum and minimum errors respectively for each sample.